\documentclass[review, number,3p]{elsarticle}

\journal{arxiv}

\setcitestyle{aysep={}}

\usepackage[hidelinks]{hyperref}

\usepackage{amsmath} 
\usepackage{amsfonts} 
\usepackage[ruled,vlined]{algorithm2e}
\usepackage{babel}  
\usepackage{wrapfig}
\usepackage{tabularx}
    \newcolumntype{L}{>{\raggedright\arraybackslash}X}
\usepackage{multicol}   
\usepackage[font=small,skip=0pt]{caption} 
\usepackage{arydshln} 
\usepackage{comment}

\usepackage{amsthm}

\makeatletter
\newtheorem*{rep@theorem}{\rep@title}
\newcommand{\newreptheorem}[2]{%
\newenvironment{rep#1}[1]{%
 \def\rep@title{#2 \ref{##1}}%
 \begin{rep@theorem}}%
 {\end{rep@theorem}}}
\makeatother

\newreptheorem{proposition}{Proposition}

\newcommand{\custcitealp}[1]{\cite{#1}}

\begin{document}
\begin{frontmatter}

\title{Maintaining Plasticity in Deep Continual Learning}

\author[inst1]{Shibhansh Dohare}
\ead{dohare@ualberta.ca}

\author[inst1]{J. Fernando Hernandez-Garcia}
\ead{jfhernan@ualberta.ca}

\author[inst1]{Parash Rahman}
\ead{parash@ualberta.ca}

\author[inst1,inst2]{A. Rupam Mahmood}
\ead{armahmood@ualberta.ca}

\author[inst1,inst2]{Richard S. Sutton}
\ead{rsutton@ualberta.ca}

\affiliation[inst1]{organization={Department of Computing Science, University of Alberta}
            }

\affiliation[inst2]{organization={CIFAR AI Chair, Alberta Machine Intelligence Institute}
            }

\begin{abstract}%

Modern deep-learning systems are specialized to problem settings in which training occurs once and then never again, as opposed to continual-learning settings in which training occurs continually. 
If deep-learning systems are applied in a continual learning setting, then it is well known that they may fail to remember earlier examples.
More fundamental, but less well known, is that they may also lose their ability to learn on new examples, a phenomenon called \textit{loss of plasticity}.
We provide direct demonstrations of loss of plasticity using the MNIST and ImageNet datasets repurposed for continual learning as sequences of tasks. 
In ImageNet, binary classification performance dropped from 89\% accuracy on an early task down to 77\%, about the level of a linear network, on the 2000th task. 
Loss of plasticity occurred with a wide range of deep network architectures, optimizers, activation functions, batch normalization, dropout, but was substantially eased by $L^2$-regularization, particularly when combined with weight perturbation. 
Further, we introduce a new algorithm---\emph{continual backpropagation}---which slightly modifies conventional backpropagation to reinitialize a small fraction of less-used units after each example and appears to maintain plasticity indefinitely.
\footnote{Code is available at \href{https://github.com/shibhansh/loss-of-plasticity}{https://github.com/shibhansh/loss-of-plasticity}}

\end{abstract}


\end{frontmatter}

\section{Loss of Plasticity}

Modern deep-learning systems have become specialized to problem settings in which training occurs once on a large data set and then never again.
Such \textit{train-once} settings were used in all the early successes of deep learning in speech recognition \cite{hinton2012speech} and image classification \cite{krishevsky2012}. 
Later, when deep learning was adapted to reinforcement learning (e.g., Mnih et al.\ \cite{mnih2015human}), techniques such as replay buffers and batching were introduced to make it very nearly a train-once setting.
Recent applications of deep learning such as GPT-3 \cite{brown2020} and DallE \cite{dallE} were also trained on data as a single large batch.
Of course, in many applications the data distribution changes over time and training must continue in some form;
the most common strategy in these cases has been to continually collect data and then, on occasion, train a new network from scratch, again in a train-once setting.
The train-once setting has been integral to the design and practice of modern deep-learning methods.

In contrast, the \textit{continual learning} problem setting focuses on learning continually from new data.
The continual learning setting is advantageous for problems where the learning system faces a changing data stream.
For example, consider a robot tasked with navigating a house.
Under the train-once setting, the robot would have to be retrained from scratch or risk becoming obsolete whenever the house's layout changed.
If the layout changed frequently, then constant retraining from scratch would be required.
Under the continual learning setting, on the other hand, the robot could simply learn from the new data and keep adapting to the changes in the house.
Continual learning has drawn increasing attention in recent years, and new specialized meetings are being organized to focus on it, such as the Conference on Life-long Learning Agents (CoLLAS). 
In this paper, we focus on the continual learning setting.

Deep learning systems, when exposed to new data, tend to forget most of what they have previously learned, a phenomenon known as ``catastrophic forgetting.''
In other words, deep learning methods do not maintain stability in continual learning problems.
This phenomenon was first shown in early neural networks in the late 1900s \cite{mccloskey1989, french1999}.
Recently, with the advent of deep learning, catastrophic forgetting has received renewed attention as many papers have been dedicated to maintaining stability in deep continual learning \cite{kirkpatrick2017, yoon2018, aljundi2019, golkar2019, riemer2019, rajasegaran2019, javed2019, aspects-of-cl}.

Different from catastrophic forgetting, and even more fundamental to continual learning, is the ability to keep learning from new data.
We refer to this ability as \emph{plasticity}.
Maintaining plasticity is essential for continual learning systems as it allows them to adapt to changes in their data streams. 
Continual learning systems that lose plasticity can become useless if their data stream changes.
In this paper, we focus on the issue of loss of plasticity.
This focus is different but complementary to the more-common focus on catastrophic forgetting. 

Do deep-learning systems lose plasticity in continual learning problems?
Some evidence that they do comes from the psychology literature of the early 2000s. Ellis and Lambon Ralph
\cite{ellis2000age}, Zevin and Seidenberg, \cite{zevin2002}, and Bonin et al.\ \cite{bonin2004} showed loss of plasticity in early neural networks in supervised regression problems. 
These papers used a setup where a set of examples was presented to the network for a certain number of epochs, and  
then the training set was expanded with additional examples and training proceeded for another number of epochs.
They found that the error for the examples in the original training set was lower than for the later-added examples after controlling for the number of epochs.
These papers provided evidence that deep learning and the backpropagation algorithm \cite{rumelhart1986learning} on which it is founded tend to produce loss of plasticity.
A limitation of these early works is that the networks used were relatively shallow by today's standards, and the algorithms used are not those that are most popular today.
For example, they did not use the ReLU activation function or the Adam optimizer.
The early psychology research on artificial neural networks provides tantalizing hints, but does not provide a clear answer to the question of whether or not modern deep learning networks exhibit loss of plasticity.

Turning now to the machine learning literature, we can find several recent studies suggesting loss of plasticity in deep learning.
Chaudhry et al.\ \cite{intransigence2018} observed loss of plasticity in continual image classification problems.
Their experiments are not fully satisfactory as a demonstration of loss of plasticity because they introduced other confounding variables.
In their setup, when a new task was presented, new outputs, called \emph{heads}, were added to the network; the number of outputs grew as additional tasks were encountered. 
Loss of plasticity was thereby confounded with the effects of interference from old heads.
Chaudhry et al.\ found that when old heads were removed at the start of a new task, the loss of plasticity was minimal, which suggests that the loss of plasticity they observed was mainly due to interference from old heads.
A second limitation of Chaudhry et al.'s study is that it\ used only ten tasks and thus did not assess loss of plasticity when deep learning methods face a long sequence of tasks.

Ash and Adams \cite{WarmStart2020} showed that initializing a neural network by first training it on a fraction of the dataset can lead to much lower final performance compared to the case in which the network is trained once on the entire dataset.
The worse performance after pretraining is an important example of loss of plasticity in deep learning and hints at the possibility of a major weakness in full continual learning with many task changes or non-stationarity. 
Berariu et al.\ \cite{plasticitystudy2021} extended Ash and Adams' work to show that the more pretraining stages there were, the worse the final performance. 
However, Berariu et al.’s experiments were performed in a stationary problem setting, and the loss of plasticity they observed was small.
Although the results in these papers hint at a fundamental loss of plasticity in deep learning systems, none directly demonstrated loss of plasticity in continual learning.

More evidence for loss of plasticity in modern deep learning has been obtained in the reinforcement learning community, where recent papers have shown substantial loss of plasticity.
Nishikin et al.\ \cite{primacy2022} showed that early learning in reinforcement learning problems could harm later learning, an effect they termed ``primacy bias."
This result could be due to the loss of plasticity of deep learning networks in continual learning settings, as reinforcement learning is inherently continual due to changes in the policy.
Lyle et al.\ \cite{capacityloss2022} also showed that some deep reinforcement learning agents could lose the ability to learn new functions over time.
These are important data points, but the conclusions that can be drawn from them are limited by the inherent complexity of modern deep reinforcement learning.
All of these papers, both from the psychology literature at the turn of the century and recent papers in machine learning and reinforcement learning, are evidence that deep-learning systems lose plasticity, but fall short of a fully satisfactory demonstration of the phenomena.

In this paper, we attempt a more definitive answer to the question of loss of plasticity in modern deep learning.
We show that deep-learning methods suffer from loss of plasticity in continual supervised learning problems and that such a loss of plasticity can be severe.
First, we demonstrate that deep learning suffers from loss of plasticity in a continual supervised-learning problem based on the ImageNet dataset that involves thousands of learning tasks. 
Using supervised learning tasks instead of reinforcement learning removes the complexity and associated confounding that inevitably arises in reinforcement learning.
And having thousands of tasks allows us to assess the full extent of the loss of plasticity.
Then we use two computationally cheaper problems (a variant of MNIST and the Slowly-Changing Regression problem) to establish the generality of deep learning's loss of plasticity over a wide range of hyperparameters, optimizers, network sizes, and activation functions. 

After showing the severity and generality of loss of plasticity in deep learning, we seek a deeper under-standing of its causes. 
This leads us to explore existing methods that might ameliorate loss of plasticity. 
We find that some methods, like Adam \cite{kingma2015} and dropout \cite{Hinton2012Dropout1}, significantly exacerbate loss of plasticity, while other methods, like $L^2$ regularization \cite[pp.~227--230]{Goodfellow-et-al-2016} and Shrink and Perturb \cite{WarmStart2020}, substantially ease the loss of plasticity in many cases.
Finally, we propose a new algorithm---continual backpropagation---that robustly solves the problem of loss of plasticity in both our systematic experiments with supervised learning problems and in a preliminary experiment with a reinforcement learning problem. 
The usual backpropagation algorithm has two crucial components: 1) initialization of the network's connection weights with small random numbers and 2) stochastic gradient descent on every presentation of a training example \cite{rumelhart1986learning}. 
Continual backpropagation extends the initialization to all time steps, by selectively reinitializing a small fraction of low-utility hidden units at every presentation of a training example.
Continual back-propagation performs both gradient descent and selective reinitialization continually.

\section{Loss of Plasticity in ImageNet}
\label{section:loss_of_plasticity_imagenet}

The main results of this paper are contained in this section.
Recall that the primary point of this paper is to establish in a more definitive way that deep learning methods lose the ability to learn in continual learning. 
To establish this, we use the classical test problem known as ImageNet and adapt it to continual learning.
ImageNet has been historically important for deep learning and is still a widely used test bed.
A classical deep learning problem like ImageNet is the best place to show the issue of loss of plasticity.
In this section, we show that deep learning suffers from loss of plasticity in a continual learning problem based on ImageNet.

Imagenet is a large database of images and their labels that has been influential throughout machine learning \cite[Image-net.org]{deng2009}
and that played a pivotal role in the rise of deep learning 
\cite{krishevsky2012}.
ImageNet allowed researchers to conclusively show that deep learning can solve a problem like object recognition, a hallmark of intelligence.
Classical machine learning methods had a classification error rate of 25\% \cite{russakovsky2015imagenet} when
deep-learning methods reduced it to below 5\% \cite{hu2018squeeze}, the human error rate on the dataset.

The ImageNet database comprises millions of images labelled by nouns (classes) such as types of animals and everyday objects. 
In the standard train-once setting, the dataset is partitioned into a training set and a test set. 
The learning system is first trained on the training set, and then the learner is frozen, and its performance is measured on the test set. 

\begin{table}[b]
    \small
    \renewcommand{\arraystretch}{0.9} 
    \centering
    \begin{tabular}{|l:c|l:c|}
    \hline
    \multicolumn{4}{|c|}{\textbf{Network Architecture for Continual ImageNet}} \\
    \hline
    \multicolumn{4}{|l|}{\textbf{Layer 1:} Convolutional + Max-Pooling} \\
    \hline
    Number of Filters   & 32    &   
    Activation          & ReLU \\
    \hdashline
    Convolutional Filter Shape        & (5,5) &   
    Convolutional Filter Stride       & (1,1)   \\ 
    \hdashline
    Max-Pooling Filter Shape   &   (2,2) &
    Max-Pooling Filter Stride &   (1,1) \\
    \hline
    \multicolumn{4}{|l|}{\textbf{Layer 2:} Convolutional + Max-Pooling} \\
    \hline
    Number of Filters   & 64   &   
    Activation          & ReLU \\
    \hdashline
    Convolutional Filter Shape        & (3,3) &   
    Convolutional Filter Stride       & (1,1)   \\ 
    \hdashline
    Max-Pooling Filter Shape   &   (2,2) &
    Max-Pooling Filter Stride &   (1,1) \\
    \hline
    \multicolumn{4}{|l|}{\textbf{Layer 3:} Convolutional + Max-Pooling} \\
    \hline
    Number of Filters   & 128   &   
    Activation          & ReLU  \\
    \hdashline
    Convolutional Filter Shape        & (3,3) &   
    Convolutional Filter Stride       & (1,1)   \\ 
    \hdashline
    Max-Pooling Filter Shape   &   (2,2) &
    Max-Pooling Filter Stride &   (1,1) \\
    \hline
    \multicolumn{4}{|l|}{\textbf{Layer 4:} Fully Connected} \\
    \hline
    Output Size     & 128  &   
    Activation      & ReLU \\
    \hline
    \multicolumn{4}{|l|}{\textbf{Layer 5:} Fully Connected} \\
    \hline
    Output Size     & 128  &   
    Activation      & ReLU \\
    \hline
    \multicolumn{4}{|l|}{\textbf{Layer 6:} Fully Connected} \\
    \hline
    Output Size     & 2  &   
    Activation      & Linear \\
    \hline
    \end{tabular}
    \smallskip
    \caption{
    Details of the artificial neural network used for the Continual ImageNet problem. 
    The network has three convolutional layers followed by three fully connected layers.
        \label{tab:imagenet-architecture}
    }
\end{table}

We sought to adapt Imagenet from the train-once setting to the continual-learning setting while mini-mizing all other changes.
The Imagenet database we used consists of 1000 classes, each of 700 images.
We constructed from these a near-endless sequence of binary classification tasks. 
For example, the first task might be to distinguish Class1 from Class2, and the second task might be to distinguish Class3 from Class4.
Each binary classification task was handled in conventional way.
The 700 images for each class were divided into 600 images for a training set and 100 images for a test set.
On each task, the deep learning network was first trained on the training set of 1200 images, and then its classification accuracy was assessed on the test set of 200 images. 
That training consisted of multiple passes through the training set, called \emph{epochs}.
After training and testing on one task, the next task was begun based on a different pair of classes.
By selecting the pairs of tasks randomly without replacement, we generated a sequence of 2000 different tasks.
All tasks used the downsampled 32x32 version of the ImageNet dataset, as is often done to save computation \cite{ImagenetDownsampled}. 
We call the resulting continual-learning problem \textit{Continual ImageNet}.
In this section, we use Continual ImageNet to establish loss of plasticity.

To show that deep learning loses the ability to learn, we applied a wide variety of standard deep learning networks to Continual ImageNet.
We describe here a specific, representative convolutional network with ReLU activation function.
The network had three convolutional-plus-max-pooling layers followed by three fully connected layers, as detailed in Table 1.
This network is narrower than those commonly used on the ImageNet dataset because here we are trying to discriminate only two classes at a time rather than all 1000.
The final layer consisted of just two units, the \emph{heads}, corresponding to the two classes.
At task changes, the input weights of the heads were reset to zero.
Resetting the heads in this way can be viewed as introducing new heads for the new classes.
This resetting of the output weights is not ideal for studying plasticity, as the learning system gets access to privileged information on the timing of task changes (and we do not use it in the experiments reported later in this paper).
We use it here because it is closest to the standard practice in deep continual learning, which is to introduce new heads for new classes \cite{kirkpatrick2017, yoon2018, aljundi2019, golkar2019}.

\begin{figure}[b]
  \centering
  \includegraphics[width=1.0\linewidth]{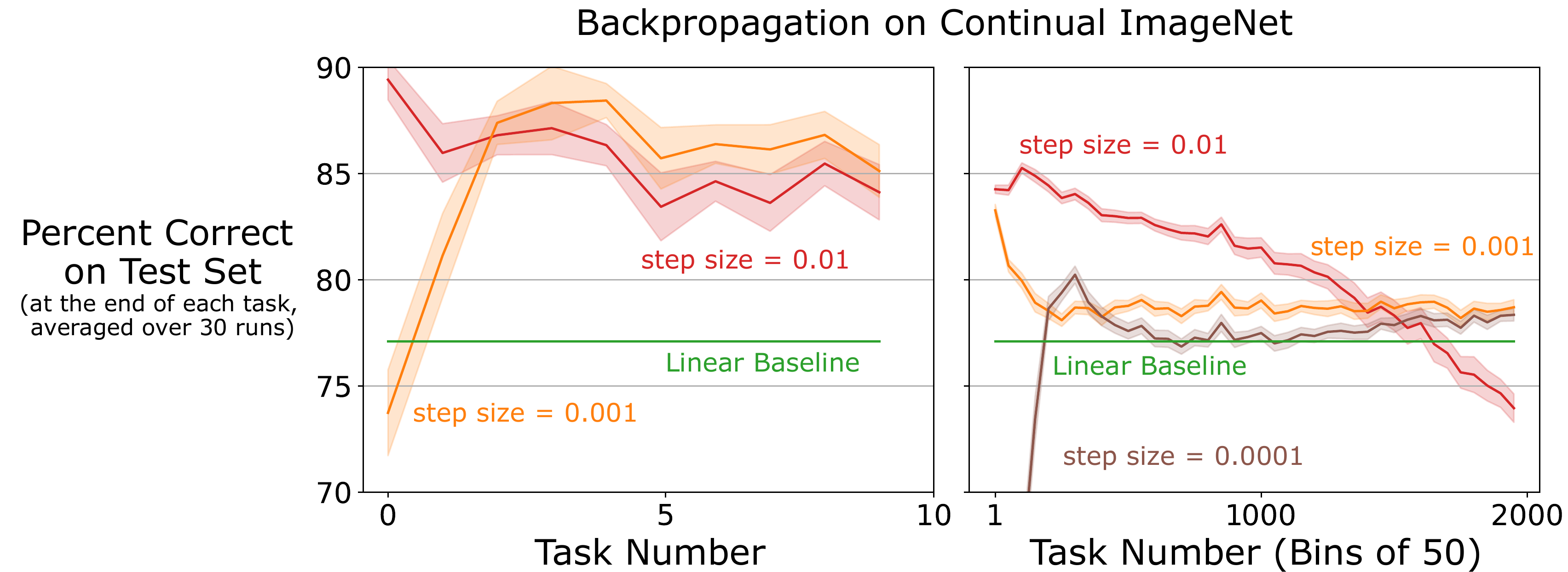}
  \vskip 0.2in
  \caption{\textbf{Loss of plasticity on a sequence of ImageNet binary classification tasks}. The first plot shows performance over the first ten tasks, which sometimes improved initially before declining. The second plot shows performance over 2000 tasks, over which the loss of plasticity was extensive. The learning algorithm was backpropagation applied in the conventional deep-learning way.}
\label{fig:imagenet_first_10_and_full}
\end{figure}

We tested many variations of the network architectures, hyperparameters, and optimizers to obtain good performance on the first binary classification task.
We now describe the details of one such training procedure applied to the network described in Table 1.
The network was trained using stochastic gradient descent (SGD) with momentum on the cross-entropy loss and initialized once prior to the first task.
For each task, the learning algorithm performed 250 passes through the training set using mini-batches of size 100. 
The momentum hyperparameter was 0.9.
We tested various step-size parameters but are only presenting the performance for step sizes 0.01, 0.001, and 0.0001 for clarity of the figure.
We performed 30 runs for each hyperparameter value, varying the sequence of tasks and other randomness.
Across different hyperparameters, the same sequences of pairs of classes were used. 

As a measure of performance on a task, we used the percentage of the 200 test images that were correctly classified.
This measure is plotted as a function of task number for the first ten tasks (left panel of Figure \ref{fig:imagenet_first_10_and_full}) and in bins of 50 tasks for all 2000 tasks (right panel of Figure \ref{fig:imagenet_first_10_and_full}) for the best step sizes.
The first data point in the right panel is the average performance on the first 50 tasks; the next is the average performance over the next 50 tasks, and so on.

The network had poor performance on the 2000th task for all hyperparameter values. 
The largest step size, 0.01, performed the best on the first two tasks, but performance then fell on subsequent tasks, eventually reaching a level below the linear baseline.
For even larger step sizes like 0.1, the network diverged after the first few tasks.
At the smaller step sizes, performance rose initially and then fell and was only slightly better than the linear baseline after 2000 tasks.
The performance for intermediate step sizes like 0.003 followed the general trend of the step sizes presented in Figure \ref{fig:imagenet_first_10_and_full}. 
We have found this to be a common pattern in our experiments:
for a well-tuned network, performance first improves, then falls substantially, ending near or below the linear baseline.
We have observed this pattern for many network architectures, hyperparameters, and optimizers.
The specific choices of network architecture, hyperparameters, and optimizers affect when the performance starts to drop, but we observed a severe performance drop for a wide range of choices.
To increase confidence in the results and to ensure that the performance drop is not due to some bug in the implementation, one of the authors independently reproduced the results of this experiment.

This experiment was designed to closely follow standard deep-learning practice in every way possible, except in requiring the network to keep learning. 
Failure to learn better than a linear network in later tasks is substantial evidence that the standard practice of deep learning simply does not work in continual problems.
We showed systematically through a direct experiment that deep learning consistently loses the ability to learn new things, i.e. deep learning loses plasticity, in continual learning problem. 
However, there are of course many other variations of deep learning that could be tried.
To reach the next level of confidence and understanding we switch in the next section to a less computationally complex problem where we can study the nuances of the phenomena more thoroughly.

\section{Robust Loss of Plasticity in Permuted MNIST}
\label{section:mnist_loss_of_plasticity}
\begin{figure}[t]
  \centering
  \includegraphics[width=\linewidth]{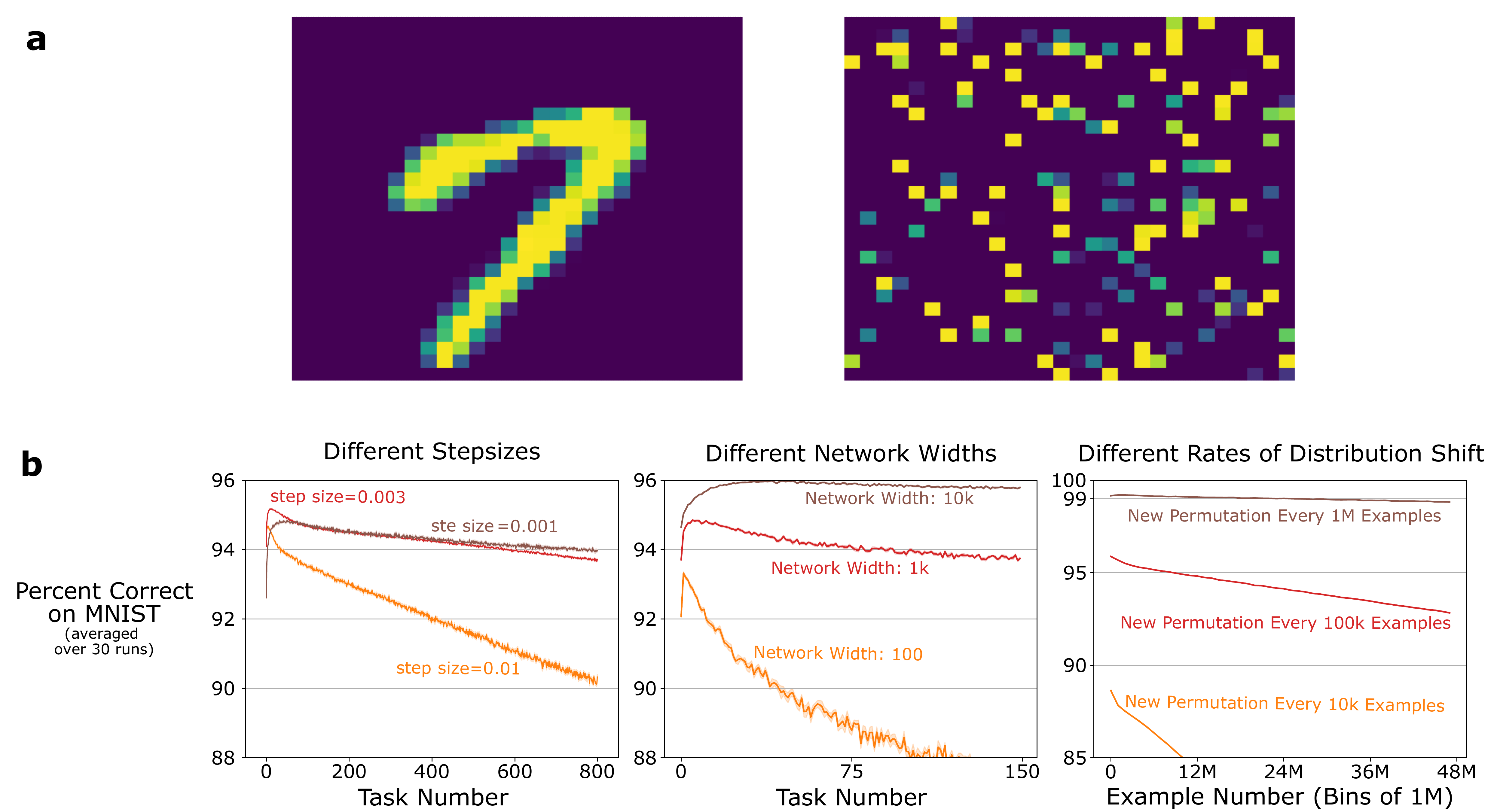}
  \vspace{5pt}
  \caption{\textbf{a:} Left: An MNIST image with the label `7'; Right: A corresponding permuted image. \textbf{b:} Loss of plasticity in Online Permuted MNIST is robust over step sizes, network sizes, and rates of change.}
\label{fig:perm-mnist}
\end{figure}

We now use a computationally cheap problem based on the MNIST dataset \cite{lecun-mnist-1998} to test the generality of loss of plasticity.
MNIST is one of the most common supervised-learning datasets used in deep learning.
It consists of 60,000 28x28 grayscale images of the handwitten digits from 0 to 9 together with their correct labels. For example, the left image in Figure~\ref{fig:perm-mnist}a shows an image that is labelled by the digit 7.
The smaller number of classes and the simpler images enables much smaller deep learning networks to perform well on this dataset than are needed on ImageNet.
The smaller networks in turn mean much less computation is needed to perform the experiments and thus experiments can be performed in greater quantities and under a variety of different conditions, enabling us to perform deeper and more extensive studies of plasticity. 
As with ImageNet, MNIST is typically used in a train-once setting.

We created a continual supervised learning problem using \textit{permuted} MNIST datasets \cite{goodfellow2014, zenke2017}.
An individual permuted MNIST dataset is created by permuting the pixels in the original MNIST dataset.
For example, the pixel that originally appeared in the upper-left corner might be moved to the lower-right corner, or to any other particular place in the image, and so on for all the other pixels.
The right image in Figure~\ref{fig:perm-mnist}a is an example of such a permuted image for an image labelled with the digit 7.
Given a way of permuting, all 60,000 images are permuted in the same way to produce the new permuted MNIST dataset.

By repeatedly randomly selecting from the approximately $10^{1930}$ possible permutations, we created a sequence of 800 permuted MNIST datasets and supervised learning tasks.
For each task, we presented each of its 60,000 images one-by-one in a random order to the learning network.
Then we moved to the next permuted MNIST task and repeated the whole procedure, and so on for up to 800 tasks.
No indication was given to the network at the time of task switching.
With the pixels being permuted in a completely unrelated way, we might expect classification performance to fall dramatically at the time of each task switch.
Nevertheless, across tasks, there could be some savings, some improvement of speed of learning, or alternatively there could be loss of plasticity---loss of the ability to learn across tasks.
The network was trained on a single pass through the data and there are no mini-batches.
We call this problem \textit{Online Permuted MNIST}.

We applied feed-forward neural networks with three hidden layers to Online Permuted MNIST. 
We did not use convolutional layers, as they could not be helpful on the permuted problem because the spatial information is lost (in MNIST, convolutional layers are often not used even on the standard, non-permuted problem).
For each example, the network estimated the probabilities of each of the 10 classes (0--9),  compared them to the correct label, and performed SGD on the cross-entropy loss.
As a measure of performance, we recorded the percentage of times the network's highest probability label was the correct one over a task's 60,000 images. We plot this per-task performance measure versus task number in Figure \ref{fig:perm-mnist}b. The weights were initialized according to a Kaiming distribution.

The first panel of Figure \ref{fig:perm-mnist}b shows the progression of performance across tasks for a network of with 2000 units per layer, and various values of the step-size parameter.
Note that performance first increased across tasks, then began falling steadily across all subsequent tasks.
This drop in performance means that the network is slowly losing the ability to learn on new tasks.
This loss of plasticiy is consistent with the loss of plasticity that we observed in ImageNet.

Next we varied the network size. Instead of 2000 units per layer, we tried 100, 1000, and 10,000  units per layer. In this experiment we ran for only 150 tasks, primarily because the largest network took much longer to run. The time courses of performance at good step sizes for each network size are shown in the middle panel of Figure \ref{fig:perm-mnist}b. The loss of plasticity with continued training is most pronounced at the smaller network sizes, but even the largest networks show some loss of plasticity. 

Next we studied the effect of the rate at which the task changed.
Going back to the original network with 2000-unit layers, instead of changing the permutation  each 60,000 examples, we now changed it after each 10,000, 100,000, or 1M examples, and ran for 48M examples total no matter how often the task changed.
The examples in these cases were selected randomly from the permuted MNIST task's dataset, with replacement. 
As a performance measure of the network on a task, we used the percent correct over all of the task's images. That is, there were only 48 data points at the slowest rate, and 4800 data points at the fastest rate.
The progression of performance is shown in the right plot in Figure \ref{fig:perm-mnist}b. 
Again performance fell across tasks, even if the change was very infrequent.
Altogether these results show that the phenomena of loss of plasticity robustly arises in this form of backpropagation.

Finally, we tested if different activation functions remove the loss of plasticity. 
We tested these activation functions in a new idealized \emph{Slowly-Changing} Regression problem. 
It is an even cheaper problem where we can run a single experiment on a CPU in 15 minutes, allowing us to do extensive studies. 
The details of this problem are given in \ref{app:scr}.
In this problem, we show loss of plasticity for networks with six different activation functions: sigmoid, tanh, ELU, leaky-ReLU, ReLU, and Swish.

So far in the paper, we performed direct tests of loss of plasticity with backpropagation.
Our experiments in Continual Imagenet, online Permuted MNIST, and Slowly changing regression show that loss of plasticity is a general phenomenon, and it can be catastrophic in some cases.
It happens for a wide range of activation functions, hyperparameters, rates of distribution change, and for both under and over parameterized net-works.
Although continual learning has been studied for a long time from \cite{caruana1997, ring1998, parisi2019}, a direct test of the ability to maintain plasticity was missing.
The experiments so far in the paper fill this gap and show that conventional backpropagation does not work in continual learning. 

There are two major goals in continual learning: maintaining stability and maintaining plasticity. 
Maintaining stability is concerned with memorizing useful information, and maintaining plasticity is about finding new useful information when the data distribution changes. 
It is difficult for current deep-learning methods to remember useful information as they tend to forget previously learned information \cite{mccloskey1989, french1999, parisi2019}.
We focused on continually finding useful information, not on remembering useful information.
Our work on loss of plasticity is different but complementary to the work on maintaining stability.

\section{Understanding Loss of Plasticity}
\label{section:understanding_loss_of_plasticity}
In this section, we turn our attention to understanding why backpropagation loses plasticity in continual learning problems.
The only difference in the learning algorithm across time are the network weights.
In the beginning, the weights were small random numbers as they were sampled from the initial distribution; however, after learning some tasks, the weights became optimized for the most recent task.
Thus, the starting weights for the next task are qualitatively different from those for the first task. 
As this difference in the weights is the only difference in the learning algorithm across time, the initial weight distribution must have some unique properties that make backpropagation plastic in the beginning. 
The initial random distribution might have many properties that enable plasticity, like the diversity of units, non-saturated units, small weight magnitude etc.

As we demonstrate now, many advantages of the initial distribution are lost concurrently with the loss of plasticity.
The loss of each of these advantages partially explains the degradation in performance that we have observed.
We then provide arguments for how the loss of these advantages could contribute to the loss of plasticity and measures that quantify how prevalent the phenomenon is.
We provide an in-depth study in the Online Permuted MNIST problem that will serve as motivation for several solution methods that could mitigate the loss of plasticity. 

The first noticeable phenomenon that occurs concurrently with the loss of plasticity is the continual increase in the fraction of constant units.
When a unit becomes constant, the gradients flowing back from the unit become zero or very close to zero, severely slowing down learning.
In the case of ReLU activations, this occurs when the output of the activations is zero for all examples of the task; such units are often said to be \textit{dead} \cite{lu2019dying, shin2020trainability}.
In the case of the sigmoid or tanh activation functions, this phenomenon occurs when the output of a unit is too close to either of the extreme values of the activation function; such units are often said to be \textit{saturated} (see \cite{glorot2010understanding} and \cite[Chapter 19]{montavon2012neural}).

To measure the number of dead units in a network with ReLU activation, we count the number of units with a value of zero for all examples in a random sample of two thousand images at the beginning of each new task.
An analogous measure in the case of sigmoid or tanh activations is the number of units that are $\epsilon$ away from either of the extreme values of the function for some small positive $\epsilon$ \cite{2015saturation}.
We only focus on ReLU networks in this section.

\begin{figure}[t]
  \centering
\includegraphics[width=\linewidth]{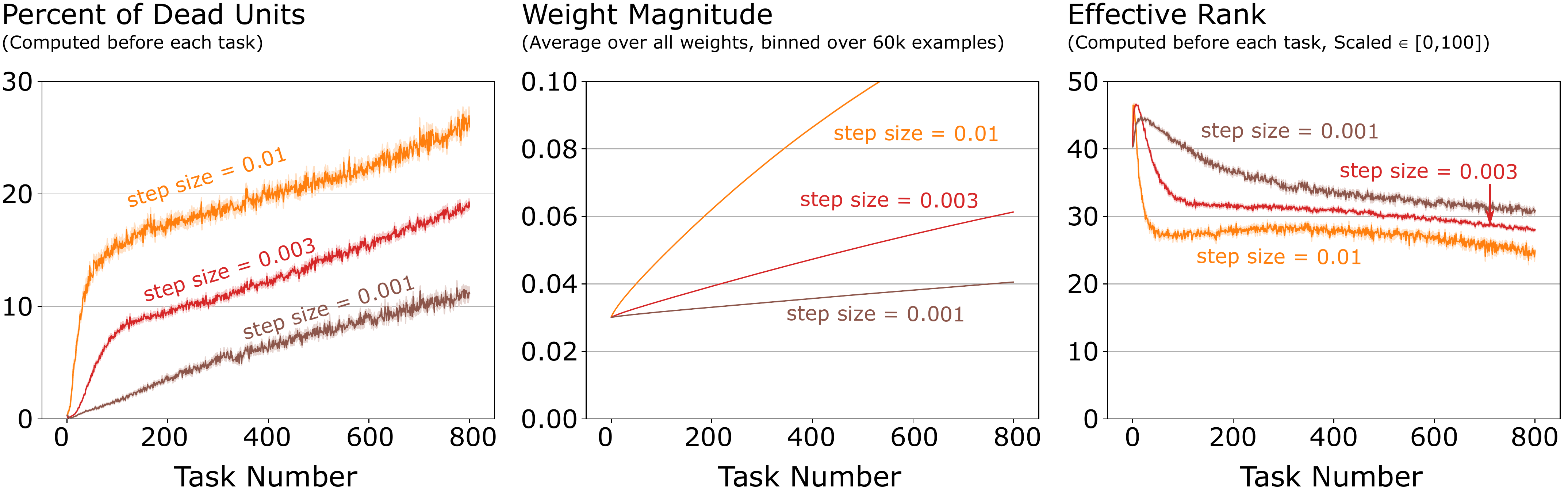}
  \caption{
  Evolution of various qualities of a deep network trained via backpropagation with different step sizes on Online Permuted MNIST for different parameter settings. 
  \textbf{Left:} Over time, the percent of dead units in the network increases for all the networks trained with backpropagation.
  \textbf{Center:} The average magnitude of the weights increases over time for all the networks trained with backpropagation.
  \textbf{Right:} The effective rank of the representation of the networks trained with backpropagation decreases over time.
  The results in these three plots are the average over 30 runs. 
  The shaded regions correspond to plus and minus one standard error.
  For some lines, the shaded region is thinner than the line width due to the standard error being small.
  \hspace{-30mm}
  }
\label{fig:bp-mnist-alternate-measures}
\end{figure}

In our experiments in the online permuted MNIST problem, the deterioration of online performance is accompanied by a large increase in the number of dead units (left panel of Figure \ref{fig:bp-mnist-alternate-measures}). 
For the step size of 0.01, up to 25\% of units die after 800 tasks.
This increase in the number of dead units partially explains why the performance of backpropagation degrades over time. 
In the next section, we will see that methods that stop the units from dying can significantly reduce the loss of plasticity. 
This suggests that the increase in dead units is one of the causes of the loss of plasticity in backpropagation.

Another phenomenon that occurs with the loss of plasticity is the steady growth of the network's average weight magnitude.
We measure the average magnitude of the weights by adding up their absolute value and dividing by the total number of weights in the network.
In the permuted MNIST experiment, the degradation of online classification accuracy of backpropagation observed in Figure \ref{fig:perm-mnist}b is associated with an increase in the average magnitude of the weights (center panel of Figure \ref{fig:bp-mnist-alternate-measures}).

The growth of the network's weights can represent a problem because large weight magnitudes are often associated with learning instability.
For example, an increase in the magnitude of the weights is associated with the well-known problem of exploding gradients in recurrent neural networks \custcitealp{pascanu2013difficulty}.
The weights of the network are a multiplicative factor of the gradient, a steady increase in the magnitude of the weights could lead to divergence of the stochastic approximation algorithms used for training the network \cite{philipp2017exploding, philipp2018gradients}.
Moreover, the convergence of descent algorithms, such as stochastic gradient descent, require gradients to remain bounded \cite{bertsekas2016}.

The last phenomenon that occurs with the loss of plasticity is the drop in the effective rank of the representation.
Similar to the rank of a matrix, which represents the number of linearly independent dimensions, the effective rank takes into consideration how each dimension influences the transformation induced by a matrix \cite{2007OlivierEffectiveRank}.
A high effective rank signals that most of the dimensions of the matrix contribute equally to the transformation induced by the matrix.
On the other hand, a low effective rank corresponds to few dimensions having any significant effect on the transformation, implying that the information in most of the dimensions is close to being redundant.

Formally, consider a matrix $\Phi \in \mathbb{R}^{n \times m}$ with singular values $\sigma_k$ for  $k=1, 2, ... , q$, and $q = \max(n, m)$.
Let $p_k = \sigma_k / \| \boldsymbol\sigma \|_1$, where $\boldsymbol\sigma$ is the vector containing all the singular values, and $\| \cdot \|_1$ is the $\ell^1$-norm.
The effective rank of matrix $\Phi$, or $\text{erank}(\Phi),$ is defined as
\begin{align}
    \text{erank}(\Phi) \overset{.}{=} \exp\left\{ H(p_1, p_2, ..., p_q)  \right\}, \text{where } 
    H(p_1, p_2, ..., p_q) = - \sum^q_{k=1} p_k \log(p_k).
\end{align}
Note that the effective rank is a continuous measure that ranges between one and the rank of matrix $\Phi$.

In the case of neural networks, the effective rank of a hidden layer measures the number of units that can produce the output of the layer.
If a hidden layer has low effective rank, then a small number of units can produce the output of the layer meaning that many of the units in the hidden layer are not providing any useful information.
We approximate the effective rank on a random sample of two thousand examples before training on each task.

In our experiments, loss of plasticity is accompanied by a decrease in the average effective rank of the network (right panel of Figure \ref{fig:bp-mnist-alternate-measures}).
This phenomenon in itself is not necessarily a problem.
After all, it has been shown that gradient-based optimization seems to favour low-rank solutions through implicit regularization of the loss function or implicit minimization of the rank itself \cite{smith2021on, 2020Razim}.
However, a low-rank solution might not be the best starting point for learning from new observations because most of the hidden units provide little to no information.

The decrease in effective rank could explain the loss of plasticity in our experiments in the following way.
After each task, the learning algorithm finds a low-rank solution for the current task, which then serves as the initialization for the next task.
As the process continues, the effective rank of the representation layer keeps decreasing after each task, limiting the number of solutions that the network is able to represent immediately at the start of each new task.

In this section, we looked deeper at the networks that lost plasticity in the Online Permuted MNIST problem. 
We noted that the only difference in the learning algorithm across time is the weights of the network, which means that the initial weight distribution has some properties that allowed the learning algorithm to be plastic in the beginning.
And as learning progressed, the weights of the network got away from the initial distribution, and the algorithm started to lose plasticity.
We found that loss of plasticity is correlated with an increasing weight magnitude, a decrease in the effective rank of the representation, and an increase in the fraction of dead units.
Each of these correlates partially explains the loss of plasticity faced by backpropagation.

\section{Existing Deep-Learning Methods for Mitigating Loss of Plasticity}

We now investigate several existing methods that could help mitigate the loss of plasticity. 
We study five existing methods: $L^2$-regularization \cite[pp.~227--230]{Goodfellow-et-al-2016}, dropout \cite{Hinton2012Dropout1}, online normalization \cite{bjork2018}, shrink-and-perturb \cite{WarmStart2020}, and Adam \cite{kingma2015}.
We chose $L^2$-regularization, dropout, normalization, and Adam because these methods are commonly used in deep learning practice.
While shrink-and-perturb is not a commonly used method, we chose it because it reduces the failure of pre-training, a problem which is an instance of loss of plasticity.
For each method, we give a brief description of how the method works and give reasons for why one expects the method to work well in general.
Then, to assess if these methods can mitigate the loss of plasticity, we apply them on Online Permuted MNIST.
We also use the three correlates of loss of plasticity found in the previous section to get a deeper understanding of the performance of these methods.
At the end of this section, we present results on Online Permuted MNIST and discuss the benefits and shortcomings of each method.

An intuitive way to address the loss of plasticity is to use parameter regularization as loss of plasticity is associated with a growth of weight magnitudes, shown in Section \ref{section:understanding_loss_of_plasticity}. 
We used $L^2$-regularization, which adds a penalty to the loss function proportional to the $\ell^2$-norm of the weights of the network. 
The networks trained using $L^2$-regularization are expected to have a smaller weight magnitude than networks trained with backpropagation alone because $L^2$-regularization directly reduces the weight magnitude.
Additionally, it is possible that stopping the growth of weight magnitudes would reduce the loss of plasticity as growing weight magnitude is associated with loss of plasticity. 

A method related to parameter regularization is shrink-and-perturb \cite{WarmStart2020}.
As the name suggests, shrink-and-perturb preforms two operations, first it shrinks all the weights and second it adds random noise to these weights.
Due to the shrinking part of shrink-and-perturb, it is expected to result in a smaller average weight magnitude than backpropagation.
Moreover, one might expect that the parameter noise introduced by shrink-and-perturb would help decrease the number of dead units and increase the effective rank of the network. 
As we observed in the previous section, loss of plasticity is associated with a large number of dead units, a low effective rank and large weights.
If shrink-and-perturb mitigates all three correlates to the loss of plasticity, shrink-and-perturb might reduce loss of plasticity.

An important technique in modern deep learning is called dropout \cite{Hinton2012Dropout1}.
Dropout randomly sets each hidden unit to zero with a small probability. 
The original motivation for dropout was to prevent hidden units from co-adapting---relying on each other to generate accurate predictions \cite{Hinton2012Dropout1}.
Moreover, dropout can also be seen as a method for introducing randomness during training, which prevents units from over-specializing and makes the overall network robust to noise \cite[pp.~255--265]{Goodfellow-et-al-2016}.
Because units in networks trained with dropout are trained not to rely on information conveyed by other units, one might expect that dropout will increase the effective rank of the hidden layers.
Aditionally, the increase in the effective rank of networks trained using dropout might also be accompanied by a decrease in the loss of plasticity.

Another commonly used technique in deep learning is batch normalization \cite{ioffe2015batch}.
In batch normalization, the output of each hidden layer is normalized and rescaled using statistics computed from each mini-batch of data.
Batch normalization was initially proposed to address the problem of internal covariate shift in neural networks \cite{ioffe2015batch}.
After its introduction, it was also found that batch normalization improved the optimization conditions when training neural networks by smoothing out the optimization landscape of the loss function \cite{santurkar2018does} and by improving the condition number, the ratio of the largest and smallest singular values, of the weight matrices of the network \cite{bjork2018}.
Normalization is expected to mitigate the problem of dead units because it is designed to ensure that units have a mean preactivation of zero and a variance of one.
Moreover, because it also improves the condition number of the weight matrices of the network \cite{bjork2018}, one might expect that networks trained with normalization would have a higher effective rank than networks trained with backpropagation alone.
As a consequence of all these effects, networks trained with normalization might show a reduced loss of plasticity than backpropagation.

No assessment of alternative methods can be complete without Adam \cite{kingma2015} as it is considered one of the most useful tools in modern deep learning.
Adam optimizer is a variant of stochastic gradient descent that uses an estimate of the first moment of the gradient scaled inversely by the second moment of the gradient to update the weights instead of directly using the gradient.
Adam has been observed to work well with non-stationary losses \cite{kingma2015}, especially in deep reinforcement learning settings \cite{lillicrap2016, mnih2016, schulman2017}.
It is possible that the robustness of Adam to non-stationary losses might help mitigate the loss of plasticity observed in non-stationary continual learning problems, such as the Permuted MNIST problem.

Properly implementing these methods for the Online Permuted MNIST problem require us to understand some details of these methods as well as the hyperparameters they come with.  $L^2$-regularization adds a penalty to the loss function proportional to the $\ell^2$-norm of the weights of the network. 
This introduces a hyperparameter $\lambda$ that modulates the contribution of the penalty term.
The incremental version of shrink-and-perturb also introduces the same regularization term as $L^2$-regularization, and it also adds a small amount of random noise to the weights on each update. 
The introduction of noise introduces another hyperparameter, the variance of the noise.
In dropout, the probability with which hidden units are set to zero is a hyperparameter, and we refer to this probability as $p$.
Batch normalization is not amenable to the online setting used in the Online Permuted MNIST problem.
Thus, we used online normalization \cite{chiley2019online}, an online variant of batch normalization.
Online normalization introduces two hyperparameters used for the incremental estimation of the statistics in the normalization steps.
Adam has two hyperparameters which are used for the incremental estimation of the statistics in the normalization steps.
These hyperparameters are used to compute the moving averages of the first and second moments of the gradient.
We used the default values of these hyperparameters proposed in the original paper.

\begin{figure}
  \centering
\includegraphics[width=\linewidth]{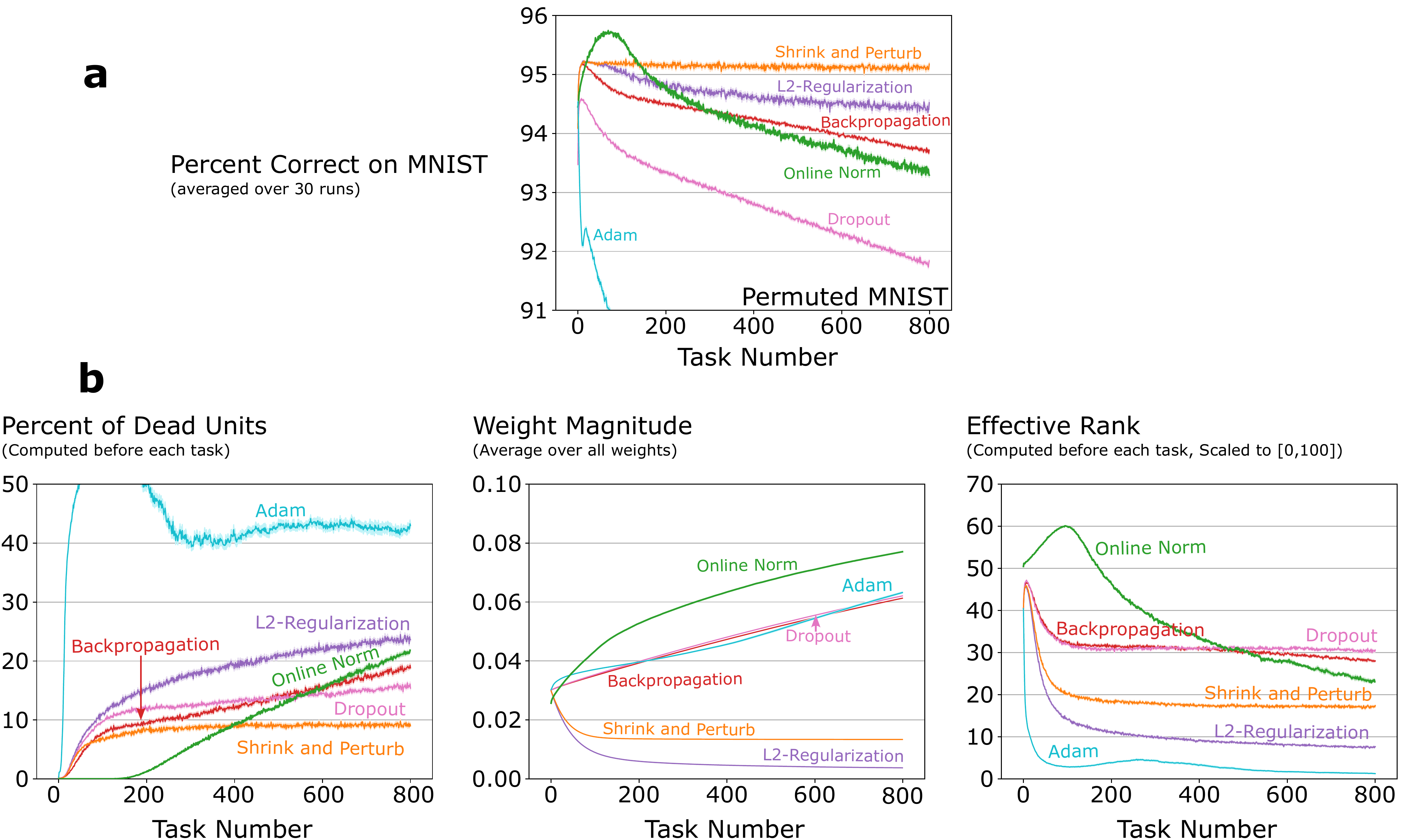}
\vspace{3pt}
\caption{\textbf{a:} Online classification accuracy of various algorithms on Online Permuted MNIST. Only $L^2$-Regularization and shrink-and-perturb have higher accuracy than backpropagation after learning 800 tasks. And, shrink-and-perturb has almost no drop in online classification accuracy over time. The results correspond to the average over 30 independent runs. The shaded regions correspond to plus and minus one standard error. \textbf{b:} Evolution of various qualities of a deep network on Online Permuted MNIST for different deep-learning algorithms. \textit{Left:} The average magnitude of the weights increases over time for all methods except for $L^2$-Regularization and Shrink and Perturb (S\&P). And these are the only two methods with an explicit mechanism to stop the weights from becoming too large. \textit{Center:} Over time, the percentage of dead units increases in all methods. S\&P  keeps the number of dead units from growing too much. And surprisingly, Online Norm starts having dead units after around 200 tasks, even though it was explicitly designed to avoid the problem of dead units. \textit{Right:} The effective rank of the representation of all methods drops over time. Dropout and S\&P stop the drop in the effective rank after around 200 tasks. The results in these plots are the average over 30 runs. The shaded regions correspond to plus and minus one standard error. For some lines, the shaded region is thinner than the line width because the standard error was small.
}
\label{fig:alternate-mnist}
\end{figure}

We tested these methods in the Online Permuted MNIST problem using the same network architecture we used in Section \ref{section:understanding_loss_of_plasticity}: a fully-connected network of three hidden layers with 2000 units each and ten output units.
Similar to \ref{section:understanding_loss_of_plasticity}, we measure the online classification accuracy on all the 60k examples of the task.

All these methods come with a set of hyperparameters, and the exact performance of the methods depends on the values of these hyperparameters.
However, we can still find hyperparameter values that represent how well the method can perform.
For each method, we tested various combinations of hyperparameters. 
In this section, we only present results for one set of these hyperparameters. 
\ref{app:hyperparameter-selection} shows the results for three different combinations of hyperparameters and provides more details on hyperparameter selection.
In general, we chose the hyperparameter value that had the highest average classification accuracy over the 800 tasks.
A picture emerges when we plot the curve for these hyperparameter values, shown in Figure \ref{fig:alternate-mnist}a.
Similar to the plots in Section \ref{section:understanding_loss_of_plasticity}, the first point in the plot is the average classification accuracy over the first task, the next over the next task and so on.
The lines corresponding to each method are not the best performance of the method, but they are representative of the best performance of the methods.
It is possible to get slightly better performance for these methods by a deeper fine-tuning of these hyperparameters, nevertheless, the pattern of results and the behaviour of these methods is well summarized by Figure \ref{fig:alternate-mnist}a. 

Changing the hyperparameter setting changed the performance of each method.
For Adam, different values of the step-size parameter changed how fast the performance dropped for Adam.
With dropout, we found that the higher the dropout probability, the faster and sharper the drop in performance.
Dropout with probability of 0.01 performed the best, and its performance was almost identical to that of backpropagation.
However, Figure \ref{fig:alternate-mnist}a shows the performance for a dropout probability of 0.1 to represent how the method performs.
For online normalization, changing the hyperparameters changed when the performance of the method peaked and it also slightly changed how fast it gets to its peak performance.
The weight-decay parameter in $L^2$-regularization controlled the peak of the performance and how fast it dropped.
Shrink-and-perturb was also sensitive to the variance of the noise. 
If the noise was too high, the loss of plasticity was much more severe, and if it was too low, it did not have any effect.

We also tested how these methods affect the three correlates of loss of plasticity we identified in Section \ref{section:understanding_loss_of_plasticity}.
Figure \ref{fig:alternate-mnist}b shows the summaries of the three correlates for the loss of plasticity for the hyperparameter settings used in Figure \ref{fig:alternate-mnist}a.

For $L^2$-regularization, the weight magnitude does not continually increase.
Moreover, as expected, the non-increasing weight magnitude is associated with a lower loss of plasticity.
However, $L^2$-regularization does not fully mitigate the loss of plasticity.
We explain this result using the other two correlates for the loss of plasticity.
While $L^2$-regularization reduces the average weight magnitude of the network, it increases the percentage of dead units and decreases the effective rank.

Similar to $L^2$-regularization, shrink-and-perturb stops the weight magnitude from continually increas-ing.
Moreover, it also reduces the percentage of dead units.
However, it has a lower effective rank than backpropagation, but still higher than that of $L^2$-regularization.
Not only does shrink-and-perturb has a lower loss of plasticity than backpropagation, but it also has minimal loss of plasticity, and it has the highest classification accuracy on the 800th task among all the methods we have tested so far.

One would have thought that dropout will increase effective rank compared to backpropagation.
This is not what we found; dropout has about the same effective rank as backpropagation.
Moreover, dropout has about the same weight magnitude and number of dead units as backpropagation.
Surprisingly, dropout resulted in a higher loss of plasticity than backpropagation.
The poor performance of dropout is not explained by our three correlates of loss of plasticity.
A thorough investigation of dropout is beyond the scope of this paper, but it will be an interesting direction for future work.

Online normalization was expected to result in fewer dead units and a higher effective rank than networks trained with backpropagation.
This happens in the earlier tasks, but both measures deteriorate over time.
In the later tasks, the network trained using online normalization has a higher percentage of dead units and a lower effective rank than the network trained using backpropagation.
The online classification accuracy is consistent with these results.
Initially, online norm results in better online classification accuracy, but by later tasks, the classification accuracy of online normalization gets lower than that of backpropagation.

Due to Adam's robustness to non-stationary losses, one would have expected that Adam would result in a lower loss of plasticity than backpropagation.
This is the opposite of what happens.
Adam's loss of plasticity can be categorized as catastrophic as it plummets drastically.
Consistent with our previous results, Adam scores poorly in the three measures corresponding to the causes for the loss of plasticity. 
There is a dramatic drop in the effective rank of the network trained with Adam.
We also tested Adam with different activation functions on the Slowly-changing regression problem and found that loss of plasticity with Adam is usually worse than with SGD.

Many methods that one might have thought would help mitigate the loss of plasticity significantly worsened the loss of plasticity.
The loss of plasticity with Adam is particularly dramatic, and the network trained with Adam quickly lost almost all of its diversity, as measured by the effective rank.
This dramatic loss of plasticity of Adam is an important result for deep reinforcement learning as Adam is the default optimizer in deep reinforcement learning and reinforcement learning is inherently continual due to the ever-changing policy.
Similar to Adam, other commonly used methods like dropout and normalization worsen the loss of plasticity.
Normalization has better performance in the beginning, but later it has a sharper drop in performance than backpropagation.
In the experiment, dropout just makes the performance worse.
We saw that the higher the dropout probability, the larger the loss of plasticity.
These results mean that some of the most successful tools in the train-once setting do not transfer to continual learning and we need to focus on directly developing tools for continual learning.

None of the existing methods fully maintain plasticity.
Some popular methods like normalization, Adam, and dropout worsen the loss of plasticity.
On the other hand, $L^2$-regularization and shrink-and-perturb reduce the loss of plasticity.
Shrink-and-perturb is particularly effective, as it almost entirely mitigates the loss of plasticity.
However, even with shrink-and-perturb, there is a minimal loss of plasticity.
Additionally, both shrink-and-perturb and $L^2$-regularization are very sensitive to hyperparameter values.
They only reduce the loss of plasticity for a very small range of parameters, while for other hyperparameter values, they make the loss of plasticity worse.
This sensitivity to hyperparameters can limit the application of these methods to continual learning.
Additionally, shrink-and-perturb does not fully resolve the three correlates of loss of plasticity, it has a lower effective rank than backpropagation, and it still has a high fraction of dead units.
Shrink-and-perturb shrinks the weights and adds randomness to them at each step.
$L^2$-regularization on its own does not mitigate the loss of plasticity.
This means that both parts of shrink-and-perturb are important to reduce the loss of plasticity.
And somewhat surprisingly, continual injection of randomness is important to mitigate the loss of plasticity.

\section{Continual Backpropagation: Stochastic Gradient Descent with Selective Reinitialization} 

We now attempt to develop a new algorithm that can fully mitigate loss of plasticity in continual learning problems as well as solve all three correlates of loss of plasticity.
In the previous section, we learned that continual injection of randomness is important to reduce the loss of plasticity.
However, the continual injection of randomness in the previous section was tied to the idea of shrinking the weights.
There exists prior work \cite{mahmood2013} that proposed a more direct way of continually injecting randomness by selectively reinitializing low-utility units in the network.
But the ideas presented in that paper were not fully developed and could only be used with neural networks with a single hidden layer and a single output, so they can not be used with modern deep learning in their current form.
In this section, we fully develop the idea of selective reinitialization so it can be used with modern deep learning.
The resulting algorithm combines conventional backpropagation with selective reinitialization.
We call it \textit{continual backpropagation}.

In one sense, continual backpropagation is a simple and natural extension of the conventional back-propagation algorithm to continual learning.
The conventional backpropagation algorithm has two main parts: initialization with small random weights and gradient descent at every time step. 
This algorithm is designed for the train-once setting, where learning happens once and never again.
It only initializes the connections with small random numbers in the beginning, but continual backpropagation does so continually.
Continual backpropagation makes conventional backpropagation continual by performing similar comp-utations at all times. 
The guiding principle behind continual backpropagation is that good continual learning algorithms should do similar computations at all times.

Continual backpropagation selectively reinitializes low-utility units in the network.
Selective reinit-ialization has two steps.
The first step is to find low-utility units and the second is to reinitialize them.
Every time step, a fraction of hidden units $\rho$, called \textit{replacement-rate}, are reinitialized in every layer. 
When a new hidden unit is added, its outgoing weights are initialized to zero.
Initializing the outgoing weights as zero ensures that the newly added hidden units do not affect the already learned function.
However, initializing the outgoing weight to zero makes the new unit vulnerable to immediate reinitialized as it has zero utility.
To protect new units from immediate reinitialization, they are protected from a reinitialization for \textit{maturity threshold} $m$ number of updates.

One major limitation of prior work on selective reinitialization is that the utility measure is limited to networks with a single hidden layer and one output.
We overcome this limitation by proposing a utility measure that can be applied to arbitrary networks.
Our utility measure has two parts. 
The first part measures the contribution of the units to its consumers.
A consumer is any unit that uses the output of a given unit.
A consumer can be other hidden units or the output units of the network.
And the second part of the utility measures units' ability to adapt.

The first part of our utility measure, called the \textit{contribution utility}, is defined for each connection or weight and each unit. 
The basic intuition behind the contribution utility is that magnitude of the product of units' activation and outgoing weight gives information about how valuable this connection is to its consumers.
If a hidden unit's contribution to its consumer is small, its contribution can be overwhelmed by contributions from other hidden units.
In such a case, the hidden unit is not useful to its consumer. 
The same measure of connection utility has been proposed for the network pruning problem \cite{hu2016}. 
We define the contribution utility of a hidden unit as the sum of the utilities of all its outgoing connections. 
The contribution utility is measured as a running average of instantaneous contributions with a decay rate, $\eta$. 
In a feed-forward neural network, the contribution-utility, $c_{l, i, t}$, of the $i$th hidden unit in layer $l$ at time $t$ is updated as
\begin{align}
\label{eq:contribution-util}
 c_{l, i, t} &= \eta*c_{l, i, t-1} + (1 - \eta)* |h_{l, i, t}| *\sum_{k=1}^{n_{l+1}} |w_{l, i, k, t}|, 
\end{align}
where $h_{l, i, t}$ is the output of the $i^{th}$ hidden unit in layer $l$ at time $t$, $w_{l, i, k, t}$ is the weight connecting the $i^{th}$ unit in layer $l$ to the $k^{th}$ unit in layer $l+1$ at time $t$, $n_{l+1}$ is the number of units is layer $l+1$.

\begin{wrapfigure}{R}{0.35\linewidth}
  \centering
\includegraphics[width=\linewidth]{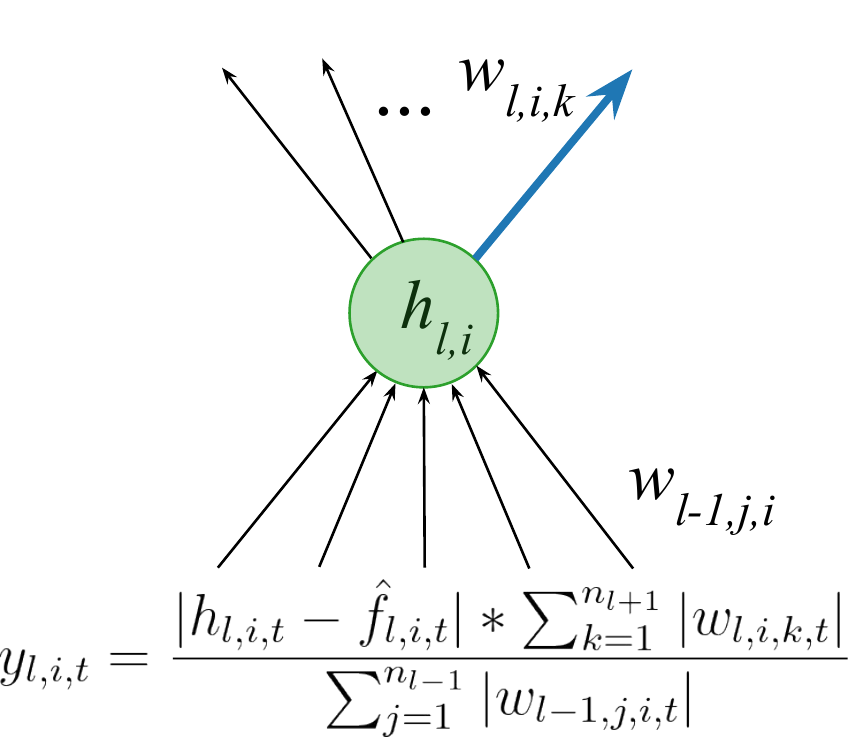}
\caption{A unit in a network. The utility of a unit at time $t$ is the product of its contribution and adaptation utilities. Adaptation utility is the inverse of the sum of the magnitude of the incoming weights. And, contribution utility is the product of the magnitude of the outgoing weights and activation ($h_{l,i}$) minus its average ($\hat{f}_{l,i}$). $\hat{f}_{l,i}$ is a running average of $h_{l,i}$.}
\label{fig:util}
\end{wrapfigure}

We use a mean-corrected version of the contribution utility.
The contribution utility is inspired by the network pruning problem.
The network pruning problem arises in the train-once problem setting, where the connections are removed after learning is done. 
However, we are studying the continual learning setting where hidden units must be replaced while learning, so we need to consider the learning process's effect on a unit's contribution.
Specifically, we need to remove the part of the contribution from the utility which is correlated with the bias.
In the special case when a consumer has just one input unit and a bias, SGD will transfer the average part of the contribution to the bias unit over time when the consumer is removed.
We define the mean-corrected contribution utility, $z_{l, i, t}$, as the product of the magnitude of the connecting weight and the magnitude of the activation minus the average value of the activation.
\begin{align}
\label{eq:centered-util}
 f_{l, i, t} &= \eta*f_{l, i, t-1} + (1 - \eta)* h_{l, i, t}, \\
\hat{f}_{l, i, t} &= \frac{f_{l, i, t-1}}{1 - \eta^{a_{l, i, t}}}, \\
 z_{l, i, t} &= \eta*z_{l, i, t-1} + (1 - \eta)* |h_{l, i, t} - \hat{f}_{l, i, t}| *\sum_{k=1}^{n_{l+1}} |w_{l, i, k, t}|,
\end{align}
where $h_{l, i, t}$ is hidden units' output, $w_{l, i, k, t}$ is the weight connecting the hidden unit to the $k$th unit in layer $l+1$, $n_{l+1}$ is the number of units is layer $l+1$, $a_{l, i, t}$ is the age of the hidden unit at time $t$. Here, $f_{l, i, t}$ is a running average of $h_{l, i, t}$ and $\hat{f}_{l, i, t}$ is the bias-corrected estimate. 
Additionally, we also transfer the unit’s average contribution, $\hat{f}_{l, i, t} * w_{l, i, k, t}$, to the bias of its consumers when the unit is removed so the consumers are less affected by the unit’s removal.
This idea of moving the average contribution to the bias has also been used for the network pruning task \cite{bias-correction}.

The second part of our utility measure captures how fast a unit can adapt.
We measure the ability to adapt as the inverse of the average magnitude of the units' input weights, and we call it \textit{adaptation-utility}.
The adaptation utility, the inverse of the average input weight magnitude, intuitively tries to capture how fast a hidden unit can get change the function it is representing.
Additionally, the inverse of the weight magnitude is a particularly reasonable measure for the speed of adaptation for Adam-type optimizers. 
In Adam, the change in weight in a single update is either upper bounded by the step size parameter or a small multiple of the step size parameter \cite{kingma2015}. 
So, during each update, hidden units with smaller weights can have a larger relative change in the function they represent.

Finally, we define the overall utility of a hidden unit as the running average of the product of its mean-corrected contribution utility and adaptation utility. The overall utility, $\hat{u}_{l, i, t}$, becomes
\begin{align}
\label{eq:util}
 y_{l, i, t} &= \frac{|h_{l, i, t} - \hat{f}_{l, i, t}| *\sum_{k=1}^{n_{l+1}} |w_{l, i, k, t}|}{\sum_{j=1}^{n_{l-1}} |w_{l-1, j, i, t}|} \\
 u_{l, i, t} &= \eta* u_{l, i, t-1} + (1 - \eta)* y_{l, i, t}, \\
\hat{u}_{l, i, t} &= \frac{u_{l, i, t-1}}{1 - \eta^{a_{l, i, t}}}.
\end{align}
The instantaneous overall utility is depicted in Figure \ref{fig:util}.

The final algorithm combines conventional backpropagation with selective reinitialization to continually inject random hidden units from the initial distribution. 
Continual backpropagation performs a gradient-descent and selective reinitialization step at each update. 
Algorithm 1 specifies the continual backpropaga-tion algorithm for a feed-forward neural network. 
Our continual backpropagation algorithm overcomes the limitation of prior work (\cite{kaelbling1993, mahmood2013}) on selective reinitialization and makes it compatible with modern deep learning. 
Prior work had two significant limitations.
First, their algorithm was only applicable to neural networks with a single hidden layer and a single output. 
Second, it was limited to LTU activations, binary weights, and SGD. 
We overcome all of these limitations. 
Our algorithm is applicable to arbitrary feed-forward networks.
We describe how to use it with modern activations and optimizers like Adam in \ref{app:cppo}.
The name ``Continual'' backpropagation comes from an algorithmic perspective. The backpropagation algorithm, as proposed by \cite{rumelhart1986learning}, had two parts, initialization with small random numbers and gradient descent.
However, initialization only happens initially, so backpropagation is not a continual algorithm as it does not do similar computations at all times. 
On the other hand, continual backpropagation is continual as it performs similar computations at all times.

\begin{algorithm}
  \caption{Continual backpropagation (CBP) for a feed-forward network with $L$ hidden layers}
  \label{alg:CBP}
\textbf{Set:} step size $\alpha$, replacement rate $\rho$, decay rate $\eta$, and maturity threshold $m$ (e.g. $10^{-4}$, $10^{-4}$, $0.99$, and $100$) \\
\textbf{Initialize:} Initialize the weights $\textbf{w}_0 , ...,\textbf{w}_L$. Let, $\textbf{w}_l$ be sampled from a distribution $d_l$\ \\
\textbf{Initialize:} Utilities $\textbf{u}_1 , ...,\textbf{u}_L$, average activation $\textbf{f}_1, ...,\textbf{f}_l$, and ages $\textbf{a}_1, ...,\textbf{a}_L$  to 0\\
\For{each input $x_{t}$}{
    \textbf{Forward pass:} pass input through the network, get the prediction, $\hat y_t$ \\
    \textbf{Evaluate:} Receive loss $l(x_t, \hat{y}_t)$ \\
    \textbf{Backward pass:} update the weights using stochastic gradient descent \\
    \For{layer $l$ in $1:L$}{
        \textbf{Update age:} $\textbf{a}_l\: +\!= 1$ \\
        \textbf{Update unit utility:} Using Equations 4, 5, and 6\\
        \textbf{Find eligible units:} Units with age more than $m$\\
        \textbf{Units to reinitialize:} $n_l\!*\!\rho$ of eligible units with the smallest utility, let their indices be $\textbf{r}$ \\
      \textbf{Initialize input weights:} Reset the input weights $\textbf{w}_{l-1}[\textbf{r}]$ using samples from $d_{l}$\\
      \textbf{Initialize output weights:} Set $\textbf{w}_{l}[\textbf{r}]$ to zero\\
      \textbf{Initialize utility, unit activation, and age:} Set $\textbf{u}_{l, \textbf{r}, t}$, $\textbf{f}_{l, \textbf{r}, t}$, and $\textbf{a}_{l, \textbf{r}, t}$ to 0
}
}
\end{algorithm}

\begin{figure*}
  \centering
\includegraphics[width=\linewidth]{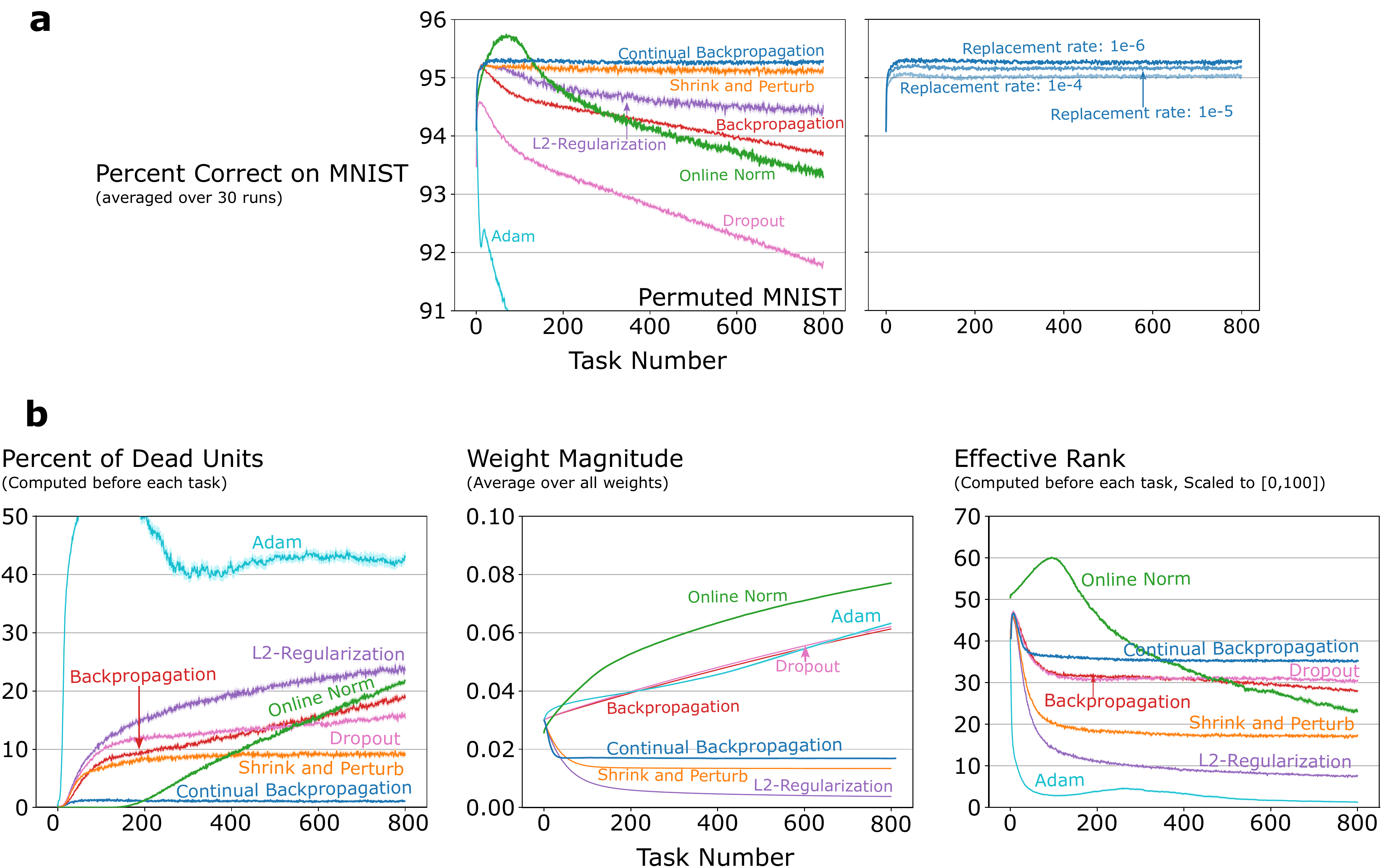}
  \caption{
  \textbf{a:} The online classification accuracy of various algorithms on Online Permuted MNIST. The performance of all algorithms except continual backpropagation degrades over time. Continual backpropagation maintains a good level of performance for a wide range of replacement rates $\rho$.
  \textbf{b:} A deeper look into various qualities of a deep network on Online Permuted MNIST using different algorithms. \textit{Left:} The average magnitude of the weights increases over time for all methods except for $L^2$-Regularization, S\&P, and continual backpropagation. And, these are the three best-performing methods. This means that small weights are important for fast learning. \textit{Center:} Over time, the percentage of dead units increases in all methods except for continual backpropagation. It has almost zero dead units throughout learning, and this happens because dead units have zero utility so they are quickly reinitialized. \textit{Right:} The effective rank of the representation of all methods drops over time. However, continual backpropagation maintains a higher effective rank than both backpropagation and shrink-and-perturb. Among all the algorithms only continual backpropagation maintains a high effective rank, low weight magnitude, and low percent of dead units.}
\label{fig:mnist-cbp}
\end{figure*}

We then applied continual backpropagation on Continual Imagenet, Online Permuted MNIST, and slowly-changing regression.
We started with Online Permuted MNIST.  
We used the same network as in the previous section, a network with 3 hidden layers with 2000 hidden units each.
We trained the network using SGD with a step size of 0.003. 
For continual backpropagation, we show the online classification accuracy for various values of replacement rates.
Replacement rate is the main hyperparameter in continual backpropagation, it controls how rapidly units are reinitialized in the network.
For example, a replacement rate of $1e-4$ for our network with 2000 hidden units in each layer would mean replacing one unit in each layer after every 5 examples.  
The hyperparameters for $L^2$-regularization, Shrink and Perturb, Online Norm, Adam, and Dropout were chosen as described in the previous section. 
The online classification accuracy of various algorithms on Online Permuted MNIST is presented in Figure \ref{fig:mnist-cbp}a. 
The results are averaged over thirty runs. 

Among all the algorithms, only continual backpropagation has a non-degrading performance.
The performance of all the other algorithms degrades over time. 
Additionally, continual backpropagation is stable for a wide range of hyperparameter values.
Note that the two best-performing algorithms are continual backpropagation and shrink-and-perturb.
And both of these algorithms enable having small weight magnitudes and diversity of representation by their design. 

Let us take a deeper look at the network that is learning via continual backpropagation.
The evolution of the correlates of loss of plasticity when using continual backpropagation is shown in Figure \ref{fig:mnist-cbp}b.
Continual backpropagation mitigates all three correlates of loss of plasticity.
It has almost no dead units, stops the network weights from growing, and maintains a high effective rank across tasks.
All algorithms that maintain a low weight magnitude reduced loss of plasticity. 
This supports our claim that low weight magnitudes are important for maintaining plasticity. 
The algorithms that maintain low weight magnitudes were continual backpropagation, $L^2$-regularization, and shrink-and-perturb. 
Shrink-and-perturb and continual backprop-agation have an additional advantage over $L^2$-regularization: they inject randomness into the network. 
This injection of randomness leads to a higher effective rank and lower number of dead units, which leads to both of these algorithms performing better than $L^2$-regularization. 
However, continual backpropagation injects randomness selectively, effectively removing all dead units from the network and leading to a higher effective rank. 
This smaller number of dead units and a higher effective rank explains the better performance of continual backpropagation.

\begin{figure}[t]
  \centering
\includegraphics[width=0.60\linewidth]{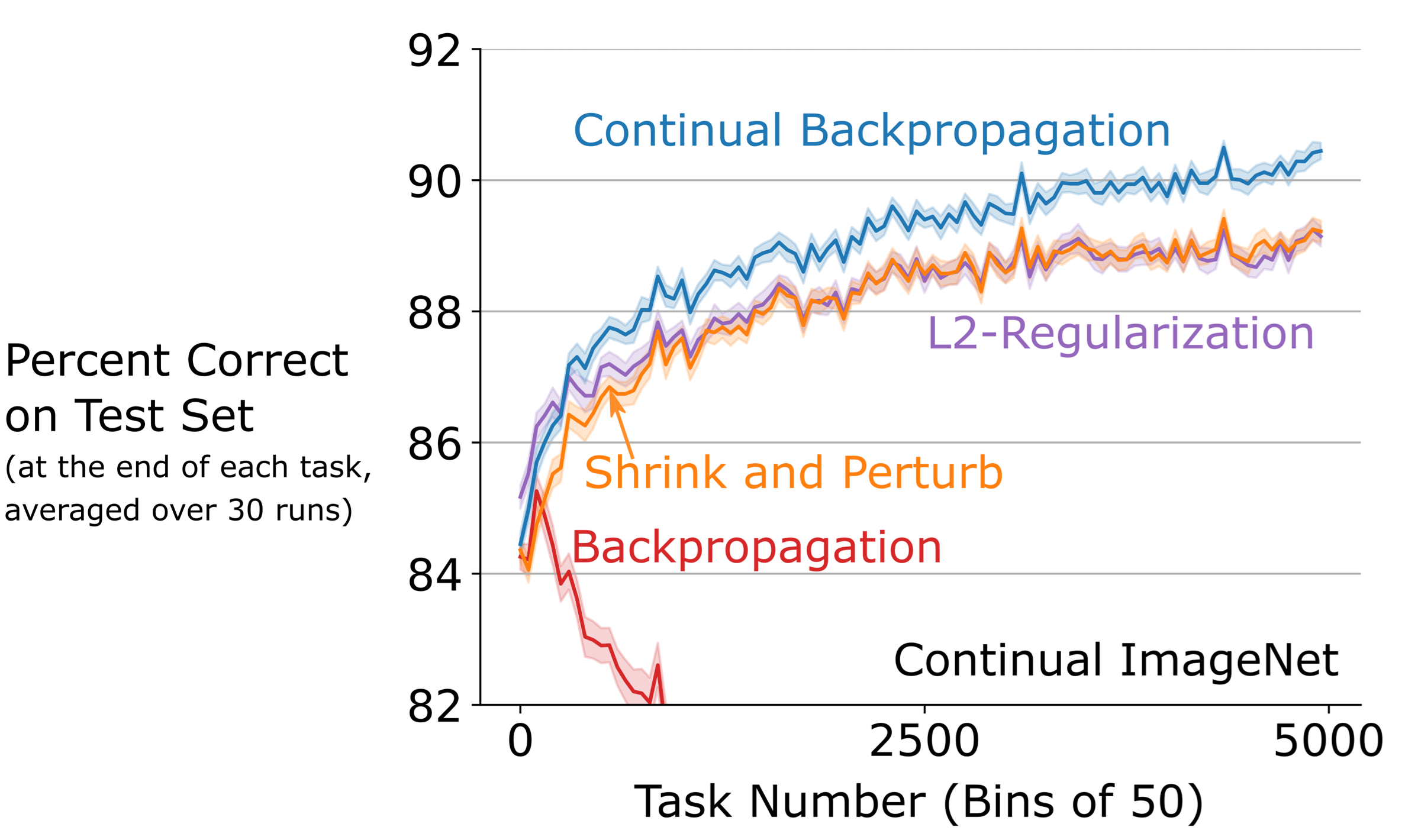}
  \caption{Continual backpropagation outperforms all the existing algorithms and fully maintains plasticity on Continual ImageNet. Its performance at the end of 5000 tasks is even better than on the first task.}
\label{fig:all-imagenet}
\end{figure}

Then we applied continual backpropagation on Continual ImageNet.
Similar to the experiments on Continual ImageNet in Section \ref{section:loss_of_plasticity_imagenet}, we used SGD with momentum and the same network that is described in Table \ref{tab:imagenet-architecture}.
We also tested L2 regularization, and shrink-and-perturb on Continual Imanget, as these are the only two methods that reduced loss of plasticity in online Permuted MNIST.
For all algorithms, we present the performance of the hyperparameter value  that had the largest average classification accuracy over 5000 tasks.
The classification accuracy of various algorithms on Continual ImageNet is shown in Figure \ref{fig:all-imagenet}.
The results are averaged over thirty runs. 
The details of the hyperparameter selection for all algorithms used in Figure \ref{fig:all-imagenet} are presented in \ref{app:hyperparameter-selection}.
The first point in the plot is the average accuracy on the first 50 tasks; the next is the average accuracy over the next 50 tasks and so on.

Continual backpropagation fully mitigates the loss of plasticity in Continual ImageNet.
Its classification accuracy on the 5000th task is better than on the first task. 
It also outperforms existing techniques like $L^2$-regularization and shrink-and-perturb.

In this section, we introduced continual backpropagation, which continually reinitializes low-utility hidden units alongside gradient descent.
Continual backpropagation fully maintains plasticity in Continual ImageNet, Online Permuted MNIST.
We also tested continual backpropagation on slowly-changing regression and found that it overcomes the loss of plasticity for all activation functions for both SGD and Adam. 
Continual backpropagation outperforms all the existing methods on all three continual learning problems.
It is much less sensitive to its hyperparameters than other algorithms like $L^2$-regularization and shrink-and-perturb. 
It also mitigates all three correlates of plasticity as it maintains a low average weight magnitude, a very small percentage of dead units, and a high effective rank.
Additionally, we performed an ablation study for the utility measure in continual backpropagation.
The results of the ablation study are present in \ref{app:ablation}, and it shows that all components of the utility measure are important for best performance.
Apart from the three supervised learning problems, we also tested these algorithms in a continual supervised learning problem.
The results of that experiment are largely consistent with the results of supervised learning experiiments and are presented in \ref{sec:crl}

Continual backpropagation opens new directions for algorithmic exploration. 
We now take a deeper look at how continual backpropagation is connected to  other ideas in the literature, and how it suggests many exciting directions for future work.
We expect that future work will explore these directions and propose new and more robust variants of continual backpropagation.

Selective reinitialization uses a utility measure to find and replace low-utility units. 
One limitation of continual backpropagation is that the utility measure is based on heuristics. 
Even though it performs well, future work on more principled utility measures will improve the foundations of continual backpropagation. 
Our current utility measure is not a global measure of utility as it does not consider how a given unit affects the overall represented function. 
One possibility is to develop utility measures where utility is propagated backwards from the loss function.
The idea of utility in continual backpropagation is closely related to connection utility in the neural network pruning literature. 
Various papers \cite{obd1989, compression2016, dst2020} have proposed different measures of connection utility for the network pruning problem. 
Adapting these utility measures to mitigate the loss of plasticity is a promising direction for new continual backpropagation algorithms. 

The idea of selective reinitialization is similar to the emerging idea of dynamic sparse training \cite{mocanu2018scalable, bellec2018deep}.
In dynamic sparse training, a sparse network is trained from scratch and connections between different units are generated and removed during training.
Removing connections requires a measure of utility, and the initialization of new connections requires a generator similar to selective reinitialization.
The main difference between dynamic sparse training and continual backpropagation is that dynamic sparse training operates on connections between units while continual backpropagation operates on units. 
Consequently, the generator in dynamic sparse training must also decide which new connections to grow.
Dynamic sparse training has achieved promising results in supervised and reinforcement learning problems \cite{chen2021chasing, sokar2021dynamic}, where dynamic sparse networks achieve performance close to dense networks even at high sparsity levels.
Dynamic sparse training is a promising idea to maintain plasticity.

The idea of adding new units to neural networks is present in the continual learning literature \cite{zhou2012, rusu2016, yoon2018}. 
This idea is usually manifested in algorithms that dynamically increase the size of the network.
So, these methods do not have an upper limit on memory requirements. 
For example, Rusu et al.\ \cite{rusu2016} presented a method that keeps expanding the size of the network as it sees more and more data. 
Their method expands the network by allocating a new sub-network whenever there is a new task. 
Although these methods are related to the ideas in continual backpropagation, none are suitable for comparison, as continual backpropagation is designed for the case when the learning system has finite memory. 
And these methods would therefore require non-trivial modification to apply to our setting of finite memory.

Prior works on the importance of initialization have focused on finding the correct weight magnitude to initialize the weights. 
Glorot and Bengio \cite{glorot2010understanding} analytically and empirically showed that it is essential to initialize the weights so that the gradients do not become exponentially small in the initial layers of a network with sigmoid activations and the gradient is preserved across layers. 
He et al.\ \cite{he2015delving} built on this idea for the ReLU activation function. 
Sutskever et al.\ \cite{sutskever2013} showed that initialization with small weights is critical for sigmoid activations as they may saturate if the weights are too large. 
Despite all this work on the importance of initialization, the fact that its benefits are only present in the beginning but not continually has been overlooked as most of these papers focused on train-once setting, where learning has to be done just once.

Loss of plasticity might also be connected to the lottery ticket hypothesis \cite{frankle2018the}. 
The hypothesis states that randomly-initialized networks contain subnetworks that can achieve performance close to that of the original network with a similar number of updates. 
These subnetworks are called winning tickets. 
We found that in continual learning problems, the effective rank of the representation at the beginning of tasks reduces over time. 
In a sense, the network obtained after training on multiple tasks has less randomness and diversity than the original random network. 
The reduced randomness might mean that the network has fewer winning tickets. 
And this reduced number of winning tickets might explain the loss of plasticity. 
Fully exploring the connection between loss of plasticity and the lottery ticket hypothesis can deepen our understanding of loss of plasticity.

Some recent works have focused on quickly adapting to the changes in the data stream \cite{finn2017, wang2017, nagabandi2019}. 
However, the problem settings in these papers were offline as they had two separate phases, one for learning and the other for evaluation. 
To use these methods online, they have to be pretrained on tasks that represent tasks the learner will encounter during the online evaluation phase. 
This requirement of having access to representative tasks in the pretraining phase is not realistic for lifelong learning systems as the real world is non-stationary, and even the distribution of tasks can change over time.
These methods are not comparable to those we studied in our work, as we studied fully online methods that do not require pretraining.

In this work, we found that methods that continually injected randomness while maintaining small weight magnitudes significantly reduced the loss of plasticity. 
Many works have found that adding noise while training neural networks can improve training and testing performance. The main benefits of adding noise have been reported to be avoiding overfitting and improving training performance \cite{holmstrom1992, gravesetal2013, neelakantan2015}. 
However, the benefits of injected noise are controversial because it can be tricky to inject noise to avoid poor performance. For example, Greff et al.\ \cite{greffetal2017} claimed adding noise always worsened their network performance. 
In our case, when the data distribution is non-stationary, we found that continually injecting noise can help maintain the plasticity in neural networks. 

The loss of plasticity also provides an alternative explanation for various recent results in deep re-inforcement learning. 
The reinforcement learning problem has various sources of non-stationarity, from the changing policy to the changing target function. 
Due to these non-stationarities, we expect that deep neural networks will suffer from a loss of plasticity in reinforcement learning problems. 
Igl et al.\ \cite{transient2021} found that deep reinforcement learning systems can lose their generalization abilities in the presence of non-stationarities. 
This loss of generalization ability could be an instance of plasticity loss, similar to the plasticity loss we observed in Continual Imagenet and online Permuted MNIST.
Kumar et al.\ \cite{kumar2021implicit} observed a reduction in the effective rank of the representation in some deep reinforcement learning algorithms.
That reduction in the effective rank is similar to the rank reduction we saw in the online Permuted MNIST problem, and their observation could be another instance of plasticity loss in deep networks.
Nikishin et al.\ \cite{primacy2022} proposed resetting some layers of the network to overcome plasticity loss. 
Their solution method, which resets some layers of the network, is in line with our results that initialization with small random numbers is essential for learning. 
Their method drastically changes the network by reinitializing a large part of the network, while continual backpropagation only slightly changes the network, but it does so continually.

\section{Discussion}

In this paper, we have shown significantly more directly, thoroughly, and systematically than in prior work that standard deep learning methods fail in continual learning settings.
We tested deep learning methods in continual supervised learning problems derived from standard datasets like Imagenet and MNIST.
These were the simplest experiments where loss of plasticity could have happened, as the experiments were designed to follow standard deep learning practices except for requiring the system to keep learning on new tasks.
Staying close to the standard practice ensured there were no confounding issues, and the experiments directly showed that deep learning methods lose plasticity.
Deep learning methods lose plasticity in both classification and regression problems. 
Plasticity loss happened for various activation functions, optimizes, network sizes, and various popular methods like dropout, normalization, Adam, and regularization, meaning that loss of plasticity is a widespread phenomenon.
We performed these experiments systematically with the highest methodological standards by performing wide parameter sweeps and at-least 30 independent runs for all methods.
By standard deep learning methods, we mean methods that are specialized to the train-once setting, and by fail, we mean that they lose the ability to learn new things.
In continual learning problems, standard deep learning methods do not work well.

We also have a significantly better understanding of loss of plasticity and the solution methods.
The root cause of loss of plasticity is that the benefits provided by initialization with small random numbers are lost over time.
A deep dive into the networks revealed that many useful properties of initialization like small weight magnitudes, few dead units, and a diversity representation are lost over time.
Many commonly used methods like Adam and dropout significantly worsened plasticity loss, while properly tuned $L^2$ regularization and shrink-and-perturb mitigated plasticity loss in many cases.
We developed the continual back-propagation algorithm, which extends the conventional backpropagation algorithm by selectively reinit-ializing a small fraction of low-utility hidden units at each update.
Continual backpropagation fully mitigated plasticity loss in all supervised continual learning problems.
It ranks units by their utility to the network’s functioning, and more improvements are possible in how ranking is done, particularly for recurrent networks.

This work opens many directions for understanding, and solving the loss of plasticity.
Although we have made significant progress in understanding the loss of plasticity, it remains unclear which specific properties of initialization with small random numbers are important for maintaining plasticity.
Recent work \cite{understanding2023, abbas2023loss} has made exciting progress in this direction, and it remains an important avenue for future work.
A crucial missing piece in the continual learning literature is the formalization of different phenomena like loss of plasticity and catastrophic forgetting. 
New papers \cite{kumar2023} are already attempting to fill this gap in the literature and future work that formalizes these concepts will be a valuable contribution to our understanding of continual learning.
One of the primary consumers of ideas in continual learning is reinforcement learning.
Recent work has improved performance in many reinforcement learning problems by applying plasticity-preserving methods \cite{nikishin2023, doro2023, sokar2023, schwarzer2023}.
Modern deep reinforcement learning methods use techniques like target networks and replay buffers to make reinforcement learning an almost train-once setting.
A more direct application of continual learning is in methods that do not use techniques to make reinforcement learning a train-once setting.
We look ahead to a promising future of better and more robust continual learning algorithms.

\section*{Acknowledgements}
We would like to thank Martha White and Prabhat Nagarajan for their feedback on the work.
We gratefully acknowledge the Digital Research Alliance of Canada for providing the computational resources to carry out the experiments in this paper. 
We also acknowledge the funding from the Canada CIFAR AI Chairs Program, DeepMind, the Alberta Machine Intelligence Institute (Amii), and the Natural Sciences and Engineering Research Council (NSERC) of Canada. 
This work was made possible by the stimulating and supportive research environment fostered by the members of the Reinforcement Learning and Artificial Intelligence (RLAI) Laboratory, particularly the insightful discussions held at the agent state group.

\newpage

\newpage

\vskip 0.2in
\bibliography{arxiv_refs}

\newpage

\appendix

\section{Methods}

\label{app:hyperparameter-selection}
In this section, we describe the procedure for selecting hyper-parameters for various algorithms for the online Permuted MNIST and Imagent problems. 

\begin{figure}[b]
  \centering
\includegraphics[width=\linewidth]{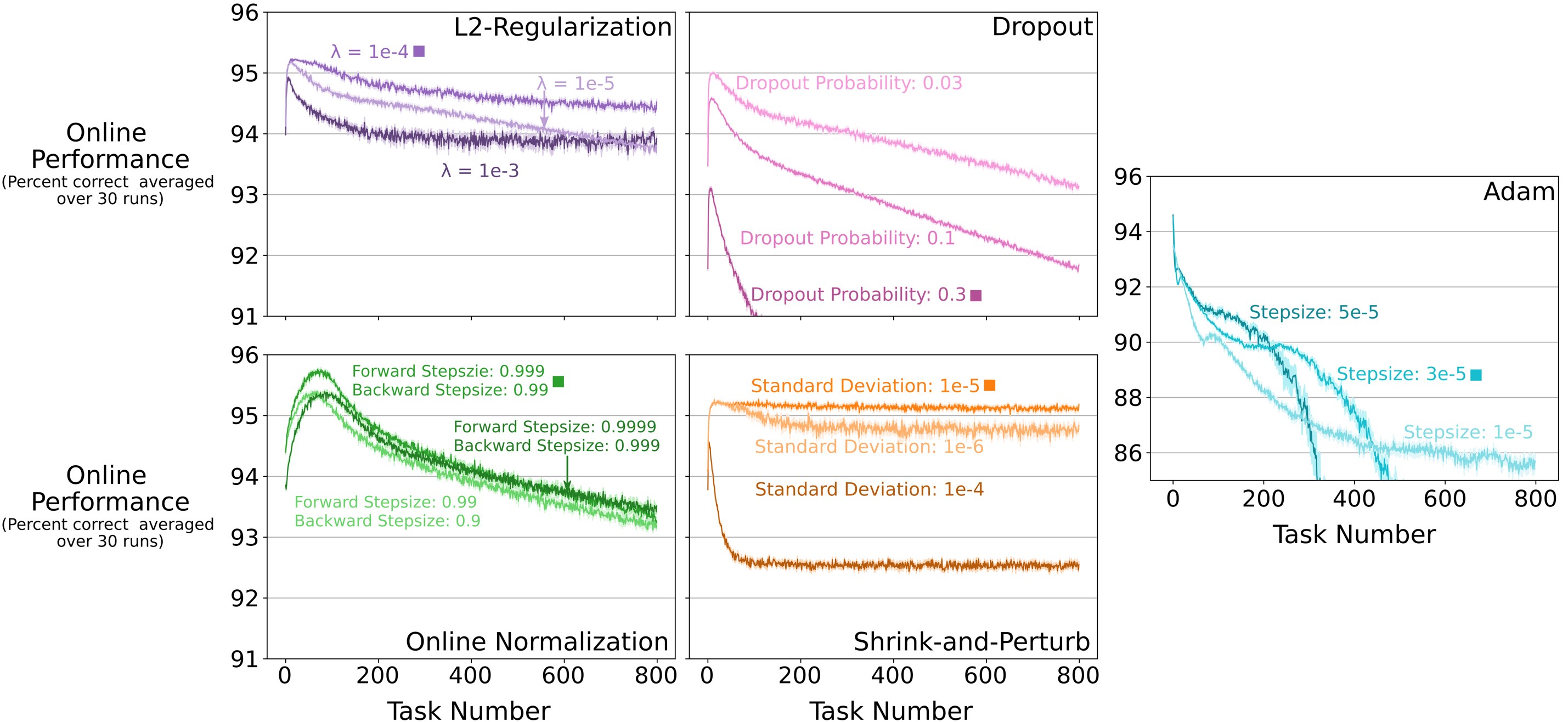}
  \caption{
  Parameter sensitivity of various algorithms on Online Permuted MNIST.
  In clockwise order and starting from the top left plot, the online classification accuracy of backpropagation with $L^2$-regularization, Dropout, Adam, Shrink-and-Perturb, and Online Normalization for various hyperparameter combinations. 
  For each method, we show three different hyperparameter settings.
  The parameter settings that were used in the main body of the paper are marked with a solid square next to their label. 
  All the lines for online norm correspond to 3 independent runs.
  All the other lines correspond to the average of 10 runs. 
  The shaded regions correspond to plus and minus one standard error.
  }
\label{fig:mnist-hyperparameter-selection}
\end{figure}

In online permuted MNIST, we tested a wide range of hyperparameters settings for each method.
In Figure \ref{fig:mnist-hyperparameter-selection} we show three different hyperparameter settings for each different algorithm.
The performance followed a similar trend for other hyperparameters values.
All the algorithms used a step size of 0.003, which was the best-performing step size for backpropagation in Figure \ref{fig:perm-mnist}b (Left).
In the case of shrink-and-perturb, the scaling factor was set to the same value as the best regularization parameter found for $L^2$-regularization, which is a suitable choice because shrink-and-perturb is equivalent to $L^2$-regularization with parameter noise.

Due to limited computational resources, we only used ten runs for the hyperparameter sweep.
For all the methods except for dropout, we selected the hyperparameter setting that resulted in the highest average percent correct during the whole training period and ran 20 more runs with that hyperparameter value.
For dropout, we selected a dropout probability of 0.1 instead of the  hyperparameter setting with the highest online classification accuracy.
We did this because the hyperparameter setting with the highest online classification accuracy was 0.03, which performs the same and is almost the same algorithm as backpropagation and would not be representative of the behaviour of dropout.

The parameter sweep for continual backpropagation is presented in the main body of the paper (\ref{fig:mnist-cbp}, right).
It follows the same experimental design except that all the lines in the plot correspond to the average over 30 runs.

We now describe the hyperparameter selection for $L^2$-regularization, shrink-and-perturb, and continual backpropagation on Continual ImageNet.
The main text presents the results for these algorithms on Continual ImageNet in Figure \ref{fig:all-imagenet}.
We performed a grid search for all algorithms to find the best set of hyperparameters.
The values of hyperparameters used for the grid search are described in Table \ref{tab:hyperparameter-imagenet}
$L^2$-regularization has two hyperparameters, step size and weight decay.
Shrink-and-perturb has three hyperparameters, step size, weight decay, and noise variance.
We swept over two hyperparameters of continual backpropagation, step size and replacement rate.
The maturity threshold in continual backpropagation was set to 100 for all sets of hyperparameters. 
For both backpropagation and $L^2$-regularization, the performance was poor for step sizes of 0.1 or 0.003.
So, we chose to only use step sizes of 0.03 and 0.01 for continual backpropagation and shrink-and-perturb.
We performed ten independent runs for all sets of hyperparameters.
We chose the set of hyperparameters with the highest average classification accuracy over 5000 tasks as the best-performing set of hyperparameters for a given algorithm.
Then we performed 20 additional runs to complete 30 runs for the best-performing set of hyperparameters to produce the results in Figure \ref{fig:all-imagenet}.

\begin{table}[]
    \centering
\begin{tabularx}{\linewidth}{|L|L|L|L|} 
    \hline
    \multicolumn{4}{|c|}{hyperparameter Selection in Continual ImageNet} \\
    \hline
    \multicolumn{1}{|L|}{Algorithm Name} & \multicolumn{1}{|L|}{values of step size} & \multicolumn{1}{|L|}{values of weight decay /replacement-rate} & \multicolumn{1}{|L|}{values of noise-variance}   \\
    \hline
    $L^2$-regularization & \{0.1, \textbf{0.03}, 0.01, 0.003\} &   \{3e-5, 1e-5, \textbf{3e-6}, 1e-6\} & Not applicable      \\ 
    \hdashline
    Shrink-and-perturb & \{0.03, \textbf{0.01}\} &   \{3e-5, \textbf{1e-5}, 3e-6, 1e-6\} & \{1e-4, \textbf{1e-5}, 1e-6, 1e-7\}      \\ 
    \hdashline
    Continual backpropagation & \{0.03, \textbf{0.01}\} & \{3e-3, 1e-3, \textbf{3e-4}, 1e-4, 3e-5\}  & Not applicable      \\ 
    \hline
    \end{tabularx}
    \caption{Values of hyperparameters used for the gird searches to find the best set of hyperparameters for all algorithms tested on Continual ImageNet. The best-performing set of values for each algorithm is boldened.}
    \label{tab:hyperparameter-imagenet}
\end{table}

\section{Loss of Plasticity With Different Activations in the Slowly Changing Regression Problem}
\label{app:scr}

There remains the issue of the network's activation function. We have been using ReLU, the most popular choice at present, but there are a number of other possibilities.
For these experiments, we switched to an even smaller, more idealized problem. 
\emph{Slowly-Changing Regression} is a computationally inexpensive problem where we can run a single experiment on a CPU core in 15 minutes, allowing us to do extensive studies. 
As its name suggests, this problem is a regression problem---meaning the labels are real numbers, with a squared loss, rather than nominal values with a cross-entropy loss---and the nonstationarity is slow and continual rather than abrupt as in a switch from one task to another.
In Slowly-Changing Regression, we study loss of plasticity for networks with six popular activation functions: sigmoid, tanh, ELU \custcitealp{clevert2016}, leaky-ReLU \custcitealp{mass2013}, ReLU \custcitealp{nair2010}, and Swish \custcitealp{ramachandran2018}.

\begin{wrapfigure}{R}{0.45\textwidth}
\begin{center}
\centerline{\includegraphics[width=0.45\textwidth]{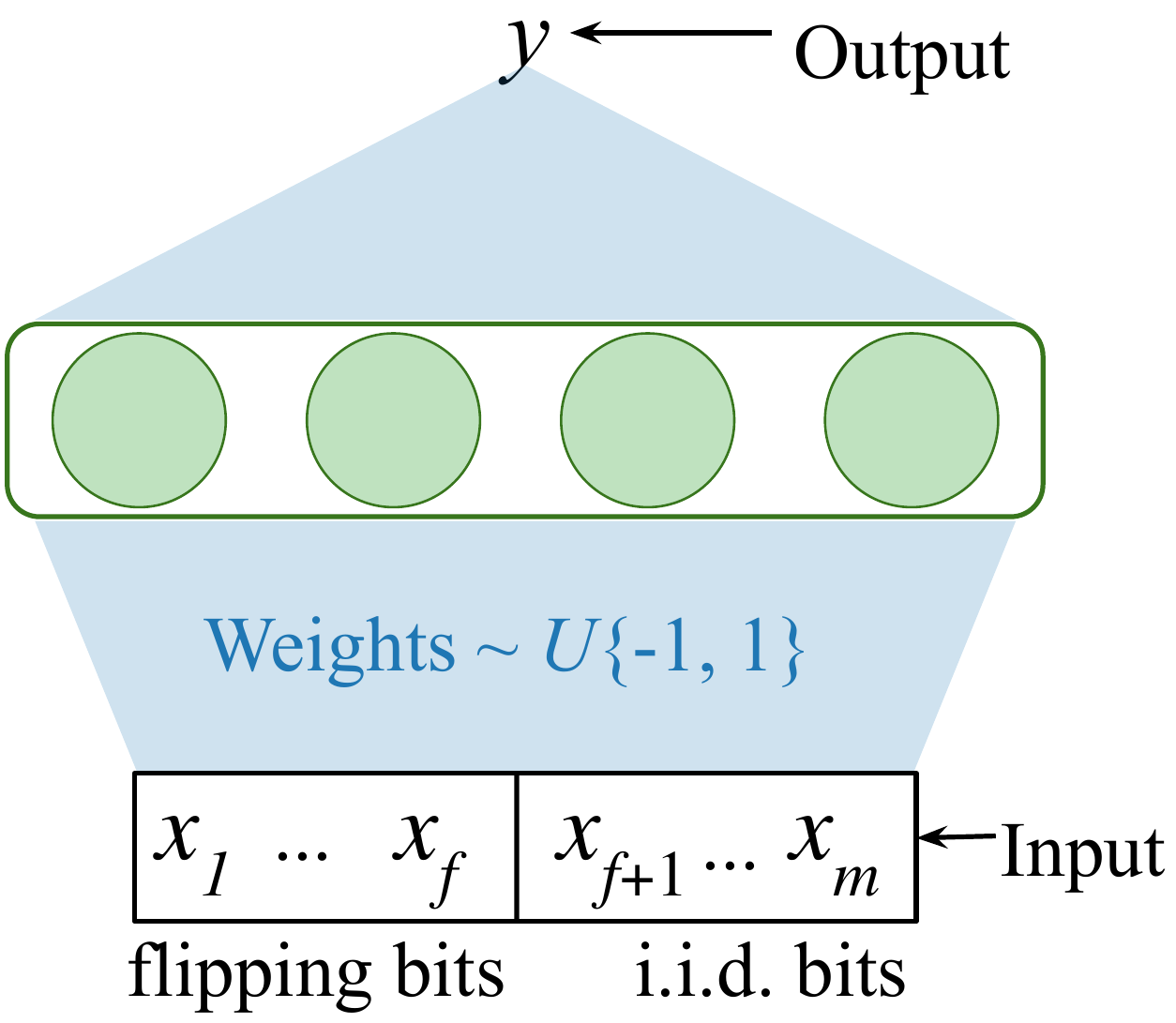}}
\end{center}
\caption{The target function and the input in the Slowly-Changing Regression problem. The input has $m+1$ bits. One of the flipping bits is chosen after every $T$ time steps, and its value is flipped. The next $m-f$ bits are i.i.d. at every time step. And, the last bit is always one. The target function is represented by a neural network with a single hidden layer of LTUs. Each weight in the target network is $-1$ or $1$.}
\label{fig:scr}
\end{wrapfigure}

In Slowly-Changing Regression, the learner receives a sequence of examples.
The input for each example is a binary vector of size $m+1$.
The input has $f$ slow-changing bits, $m-f$ random bits, and then one constant bit.
The first $f$ bits in the input vector change slowly.
After every $T$ examples, one of the first $f$ bits is chosen uniformly at random, and its value is flipped.
These first $f$ bits remain fixed for the next $T$ examples.
The parameter $T$ allows us to control the rate at which the input distribution changes.
The next $m-f$ bits are randomly sampled for each example. 
Lastly, the $(m+1)$-th bit is a bias term with a constant value 1.

The target output is generated by running input through a neural network, which is set at the start of the experiment and kept fixed.
As this network generates the target output and represents the desired solution, we call it the \emph{target network}.
The weights of the target networks are randomly chosen to be +1 or -1.
The target network has one hidden layer with a Linear Threshold Unit \cite{mcculloch1943}, or LTU, activation. 
The value of the $i$-th LTU is one if the input is above a threshold $\theta_i$, and zero otherwise. 
$\theta_i$ is set equal to $(m+1) \cdot \beta - S_i$, where $\beta \in [0,1]$ and $S_i$ is the number of input weights with negative value \cite{sutton1993}. 
The details of the input and target function in the slowly-changing regression problem are also described in Figure \ref{fig:scr}.

The details of the specific instance of the slowly-changing regression problem we use in this paper and the learning network used to predict its output are listed in Table \ref{tab:scr}. 
Note that the target network is more complex than the learning network as the target network is wider, with 100 hidden units, while the learner has just five hidden units. 
Thus, because the input distribution changes every $T$ example and the target function is more complex than what the learner can represent, there is a need to track the best approximation.
 
\begin{table}[]
    \centering
    \begin{tabular}{|l|l|c|}
    \hline
    \multicolumn{3}{|c|}{slowly-changing regression Problem Parameters} \\
    \hline
    \multicolumn{1}{|c|}{Parameter Name} & \multicolumn{1}{|c|}{Description} & Value   \\
    \hline
    $m$             & Number of input bits          &   21      \\ 
    \hdashline
    $f$             & Number of flipping bits       &   15      \\ 
    \hdashline
    $n$             & Number of hidden units        &   100      \\ 
    \hdashline
    $T$             & Duration between bit flips     &   10,000 time steps  \\
    \hdashline
    Bias            & Include bias term in input and output layers
                                                    & True\\ 
    \hdashline
    $\theta_i$      & LTU Threshold                 &   $(m+1) \cdot \beta - S_i$\\ 
    \hdashline
    $\beta$         & Proportion used in LTU Threshold 
                                                    & 0.7 \\ 
    \hline
    \multicolumn{3}{|c|}{Learning Network Parameters}\\
    \hline
    \multicolumn{2}{|c|}{Parameter Name} & Value   \\
    \hline
    \multicolumn{2}{|l|}{Number of hidden layers}               & 1   \\
    \hdashline
    \multicolumn{2}{|l|}{Number of units in each hidden layer}  & 5   \\
    \hline
    \end{tabular}
    \vspace{3mm}
    \caption{Implementation details for the slowly-changing regression problem and the learning network}
    \label{tab:scr}
\end{table}

\begin{figure}
  \centering
\includegraphics[width=\linewidth]{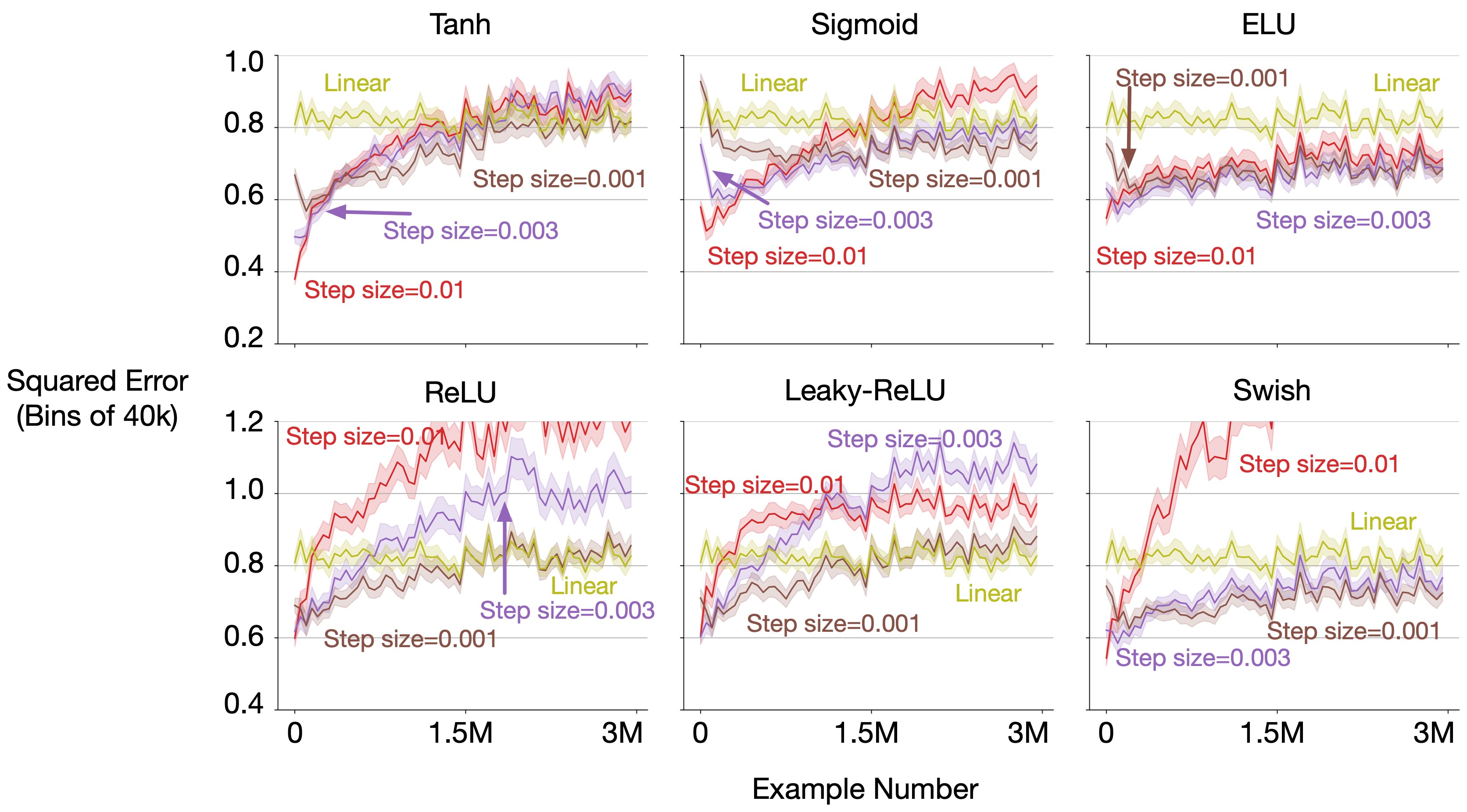}
\caption{Loss of plasticity is robust across different activations. \label{fig:bp-scr}
}
\end{figure}

We applied learning networks with different activation functions to the slowly-changing regression.
The learner used the backpropagation algorithm to learn the weights of the network.
We used uniform Kaiming distribution \cite{he2015delving} to initialize the learning network's weights. 
The distribution is $U(-b, b)$ with bound, $b=gain*\sqrt\frac{3}{num\_inputs}$, where $gain$ is chosen such that the magnitude of inputs does not change across layers. 
For tanh, Sigmoid, ReLU, and Leaky-ReLU,  $gain$ is 5/3, 1, $\sqrt{2}$, and $\sqrt{2/(1 + \alpha^2)}$ respectively. 
For ELU and Swish, we used $gain=\sqrt{2}$, as was done in the original papers \cite{clevert2016, ramachandran2018}.

We ran the experiment on the Slowly-changing regression problem for 3M examples. 
For each activation and value of step size, we performed 100 independent runs.
First, we generated 100 sequences of examples (input-output pairs) for  the 100 runs.
Then these 100 sequences of examples were used for experiments with all activations and values of the step size parameter.
We used the same sequence of examples to control the randomness in the data stream across activations and step sizes. 

The results of the experiments are shown in Figure \ref{fig:bp-scr}.
We measured the squared error, the square of the difference between the true target and the prediction made by the learning network.
In Figure \ref{fig:bp-scr}c, the squared error is presented in bins of 40k examples. 
This means that the first data point is the average squared error on the first 40k examples, the next is the average squared error on the next 40k examples, and so on.
The shaded region in the Figure shows the standard error of the binned error.

Figure \ref{fig:bp-scr} shows that in Slowly-changing regression, after performing well initially, the error increases for all step sizes and activations.
For some activations like ReLU and tanh, loss of plasticity is severe, and the error increases to the level of the linear baseline.
While for other activations like ELU, loss of plasticity is less severe, but still there is a significant loss of plasticity.
These results mean that the loss of plasticity is not resolved by using commonly used activations.

\section{Extension to a continual RL problem, Slippery Ant}
\label{sec:crl}

One of the major consumers of ideas in continual learning is the field of reinforcement learning as there are many sources of non-stationarity in reinforcement learning.
In this section, we explore if the findings from continual supervised learning transfer to continual reinforcement learning (RL).
However, a full exploration of this idea is beyond the scope of the current work.
Fully exploring the issue of loss of plasticity in reinforcement learning is much harder because many additional confounders arise in reinforcement learning, and proper empirical research will require an extensive parameter sweep over tens of hyperparameters in each algorithm.
So, the results in this section are preliminary as we did not perform full parameter sweeps in this section.

\begin{wrapfigure}{R}{0.45\textwidth}
  \centering
\centerline{\includegraphics[width=0.45\textwidth]{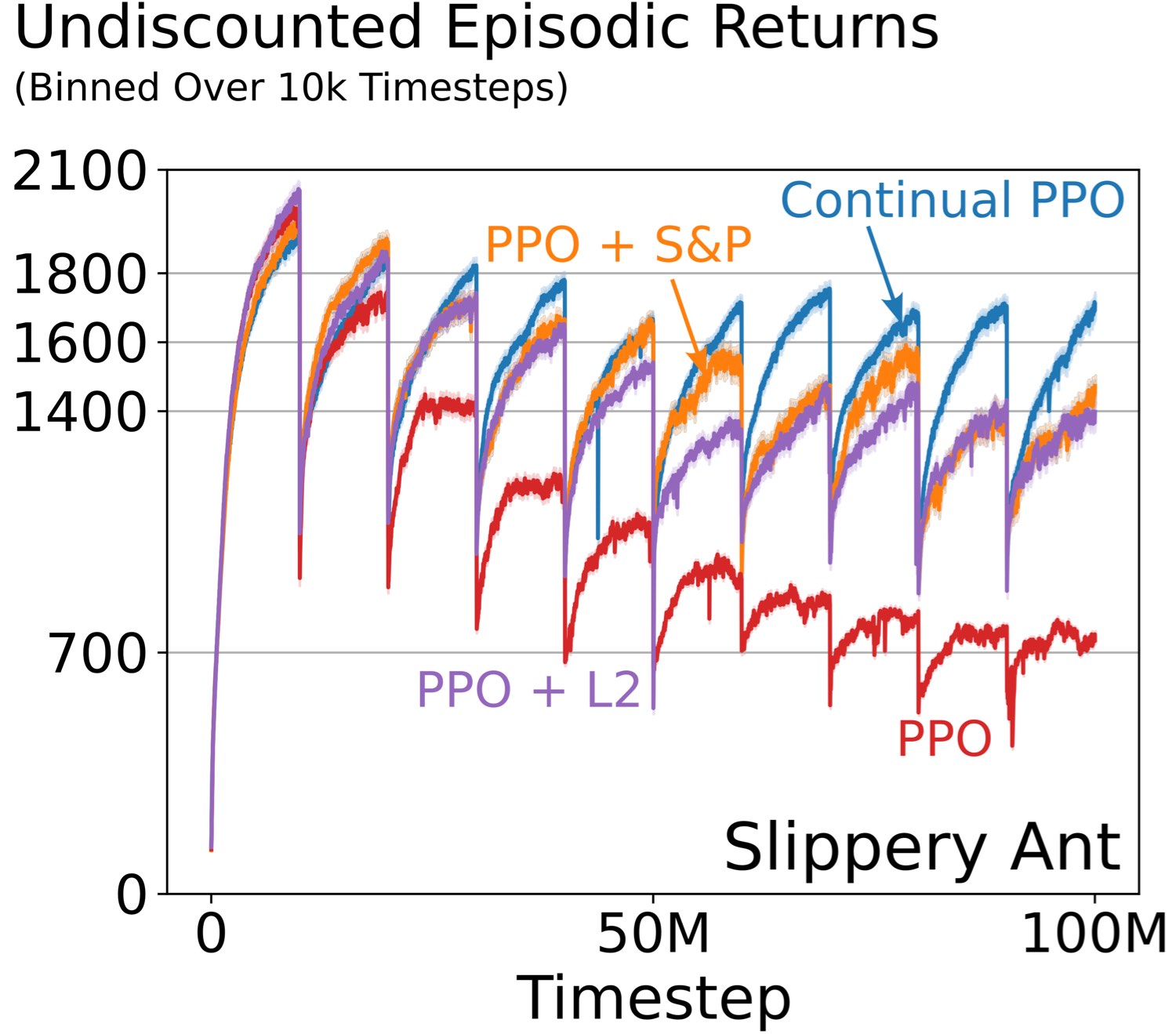}}  \caption{Performance of PPO, which is based on backpropagation, degraded rapidly on slippery Ant. In contrast, all three Continual PPO, PPO+S\&P, and PPO+$L^2$ perform substantially better than PPO. Continual PPO performed best and only had a minimal drop in performance over the 100M time steps.}
\label{fig:rl-result}
\end{wrapfigure}

We developed a new continual RL problem, Slippery-Ant.
Slippery-Ant is a continual variant of Pybullet's \cite{coumans2021} Ant problem. 
A continual variant is needed as the problems in Pybullet are stationary. 
In our problem, the environment changes after a pre-specified time, making it necessary for the learner to adapt. 
We change the friction between the agent and the ground of the standard Pybullet Ant problem. 
In the standard problem, the value of friction is $1.5$. 
After every 10M time steps, we change the friction by log-uniformly sampling it from $[10^{-4}, 10^{+4}]$.

We used the PPO algorithm \cite{schulman2017} to solve the Slippery-Ant problem. 
Two separate networks were used for the policy and the value function, and both had two hidden layers with 256 units.
These networks had \textit{tanh} activation as it performs the best with on-policy algorithms like PPO \cite{andrychowicz2021}.
The networks were trained using Adam alongside PPO to update the weights in the network.

To combine continual backpropagation with PPO, we used continual backpropagation instead of back-propagation to update the networks' weights. 
Whenever the weights are updated using Adam, we also update them using generate-and-test. 
We call the resulting algorithm \textit{Continual PPO}, and we describe it in \ref{app:cppo}.

All the algorithms that we tested are built on PPO and we used a standard set of parameters for PPO.
In addition to the parameters for PPO, all algorithms require us to choose additional parameters.
For $L^2$, we chose the best weight decay. 
For S\&P, we chose the best perturb for the value of weight decay found for $L^2$.
And for continual PPO, we chose the best replacement rate, maturity threshold pair. 
All parameter settings in are described in \ref{app:cppo}. 
All other parameters were the same in PPO, PPO+$L^2$, PPO+S\&P and continual PPO. 
The performance of PPO, PPO+$L^2$, PPO+S\&P and continual PPO for Lecun initialization \cite{lecun-1998} on Slippery Ant is shown in Figure \ref{fig:rl-result}. 
The results are averaged over 100 runs.

The performance of PPO degrades dramatically as the environment changes in Slippery Ant. 
Whenever the environment changes, there will be a drop in performance as the agent has to learn a new policy.
Due to these sudden drops the performance plot looks like a saw tooth, where each tooth shows the performance on one task.
Across tasks, the performance that the PPO agent drops dramatically.
This degradation is similar to the degradation of backpropagation's performance on continual supervised learning problems, where it performs well initially, but its performance gets worse over time. 
This similarity with the performance of backpropagation is not surprising as backpropagation lies at the foundation of modern deep reinforcement learning algorithms like PPO.

All three continual PPO, PPO+S\&P, and PPO+$L^2$ perform substantially better than PPO, and continual PPO performs best, where it continually performs almost as well as it does initially.
There is still a small performance drop with continual PPO, which could be due to the additional confounders introduced by PPO.
Fully understanding the mitigating loss of plasticity in continual RL problems is an important open avenue for future work.

In this section, we performed preliminary experiments in a reinforcement learning problem.
These preliminary results in continual reinforcement learning problems are largely consistent with those in continual supervised learning problems.
The fact that continual PPO does not fully maintain plasticity is interesting and worthy of future exploration.
Additionally, we showed that continual backpropagation can be directly combined with existing algorithms.
Continual PPO is an example of how to create continual variants of existing deep reinforcement learning algorithms.

\section{Ablation Study for the utility measure}
\label{app:ablation}

\begin{figure*}[tp]
\vskip 0.2in

\centerline{\includegraphics[width=0.7\textwidth]{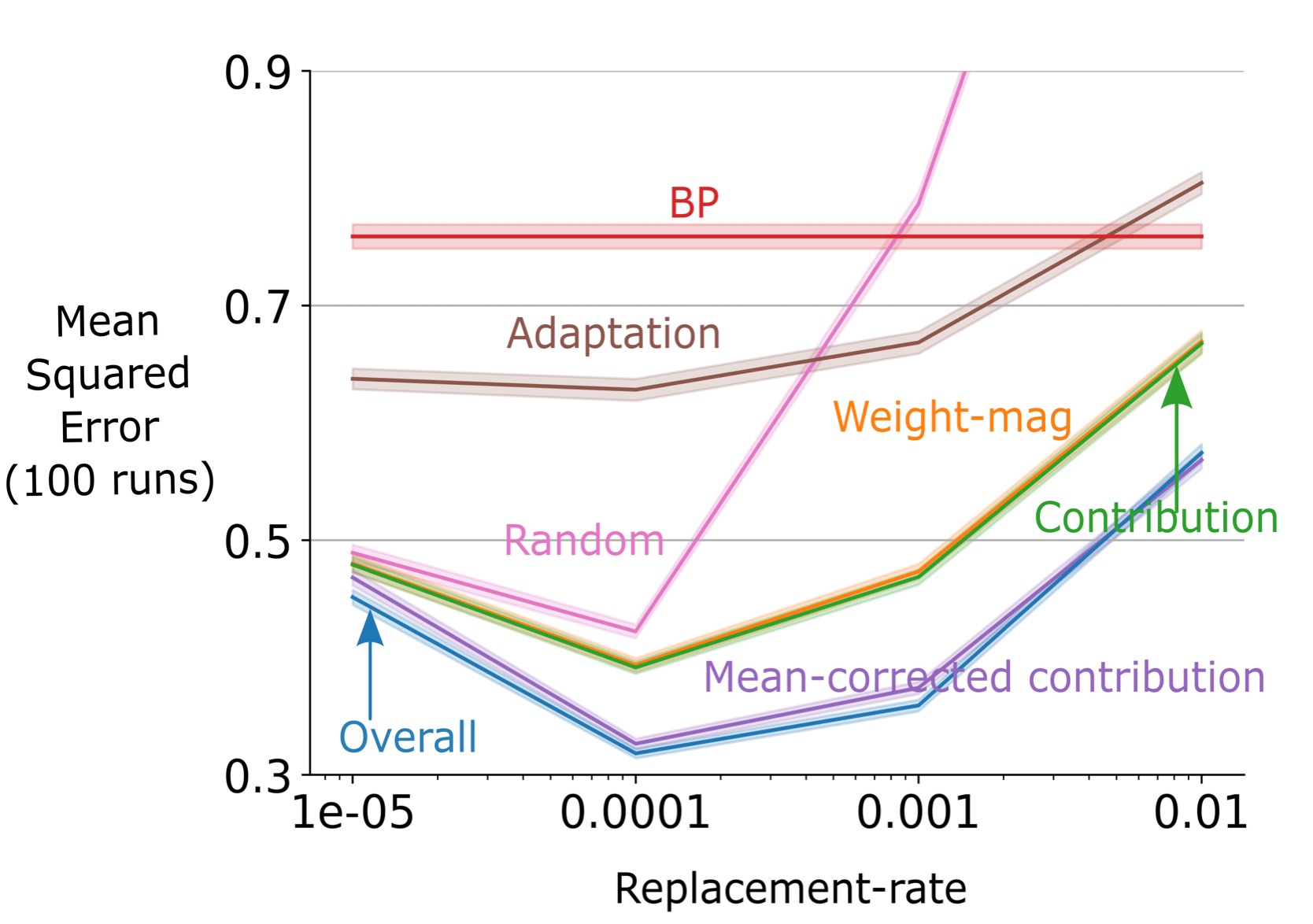}}
\caption{Parameter sweep for various utility measures on the slowly-changing regression problem. The overall utility measure performs better than all other measures of utility, including the weight-magnitude utility and random utility.}
         \label{fig:ablation-bfp}
\vskip -0.2in
\end{figure*}

The overall utility measure consists of two parts, the contribution utility and the adaptation utility. We compared various parts of the utility measure on the Slowly-changing regression problem. We use a learning network with tanh activation and Adam with a step size of 0.01. We also compare our utility measure with random utility and weight-magnitude-based utility. The results for this comparison are presented in Figure \ref{fig:ablation-bfp}. We compared the following utility measures.

\begin{itemize}
    \item Random utility: Utility, $r$ at every time step is uniformly randomly sampled from $U[0, 1]$
    \begin{equation*}
    r_{l, i, t} = rand(0,1)
    \end{equation*}
    \item Weight-magnitude based utility: $wm$ at every time step is updated as:
    \begin{equation*}
    wm_{l, i, t} =  (1 - \eta)* \sum_{k=1}^{n_{l+1}} |w_{l, i, k, t}| + \eta * wm_{l, i, t-1}
    \end{equation*}
    \item Contribution utility, $c$, at every time step is updated as described in Equation \ref{eq:contribution-util}
    \item Mean corrected contribution utility, $z$, at every time step is updated as described in Equation \ref{eq:centered-util}
    \item Adaptation utility, $a$, at every time step is updated as:
    \begin{equation*}
    a_{l, i, t} =  (1 - \eta)* \frac{1}{\sum_{j=1}^{n_{l-1}} |w_{l, j, i, t}|} + \eta*a_{l, i, t-1}
    \end{equation*}
    \item Overall utility, $u$ ,at every time step is updated as described in Equation \ref{eq:util}
\end{itemize}

The results for various utility measures are presented in Figure \ref{fig:ablation-bfp}. The results show that all the components of our utility measure are needed for the best performance. They also show that our utility measure performs significantly better than random utility and weight-magnitude utility.

\begin{figure*}[tp]
\vskip 0.2in

\centerline{\includegraphics[width=0.7\textwidth]{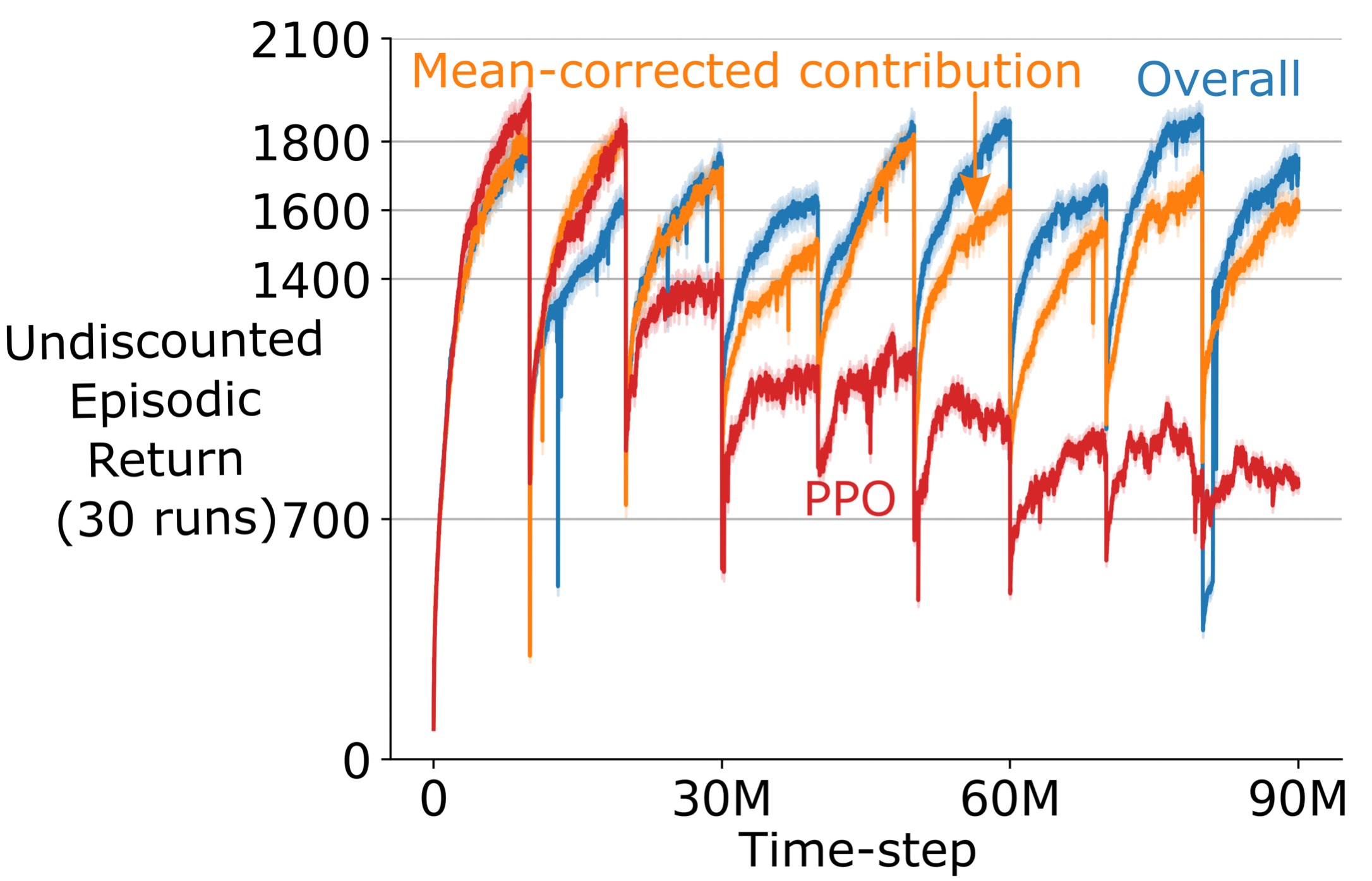}}
         \caption{Continual PPO with overall utility and mean-corrected contribution utility on Slippery Ant. Both utility measures significantly outperformed PPO and overall utility performed better than mean-corrected contribution utility.}
    \label{fig:ablation-rl}
\vskip -0.2in
\end{figure*}

Next, we compared the two best-performing utility measures from the slowly changing regression, overall and mean-corrected contribution, on Slippery Ant. The results are presented in Figure \ref{fig:ablation-rl}. The results show that both utility measures perform significantly better than PPO. But, the overall utility performs significantly better than the other utility measure. This difference is more pronounced near the end when continual PPO with the overall utility measure performs almost as well as at the beginning.

\section{Continual PPO}
\label{app:cppo}

We used the following values of the parameters for PPO, PPO+$L^2$ and Continual PPO in our experiments on our non-stationary RL problem. For Continual PPO and PPO+$L^2$ specific parameters, we chose the best-performing values.
\begin{itemize}
    \item Policy Network: (256, tanh, 256, tanh, Linear) + Standard Deviation variable
    \item Value Network (256, tanh, 256, tanh, linear)
    \item iteration size: 4096
    \item num epochs: 10
    \item mini-batch size: 128
    \item GAE, $\lambda$: 0.95
    \item Discount factor, $\gamma$: 0.99
    \item clip parameter: 0.2
    \item Optimizer: Adam
    \item Optimizer step size: $1e-4$
    \item Optimizer $\beta$s: (0.9, 0.999)
    \item weight decay (for $L^2$): $\{10^{-3}, 10^{-4}, 10^{-5}, 10^{-6}\}$
    \item replacement rate, maturity threshold (for CPPO): $\{(10^{-3}, 1e2), (10^{-3}, 5e2), (10^{-4}, 5e3)$, \\ $(10^{-4}, 5e2), (10^{-5}, 1e4), (10^{-5}, 5e4)\}$
\end{itemize}

PPO uses the Adam optimizer. 
To properly use Adam with continual backpropagation, we need to modify the Adam optimizer.
Adam maintains estimates of the average gradient and the average squared gradient. 
Continual backpropagation resets some of the weights whenever a feature is replaced. 
Whenever continual backpropagation resets a weight, we set its estimated gradient and squared gradient to zero. 
Adam also maintains a 'timestep' parameter to get an unbiased estimate of the average gradient. 
Again, whenever continual backpropagation resets a weight, we set its timestep parameter to zero.
The continual backpropagation algorithm with Adam is described in Algorithm \ref{alg:cbp-adam}.
And Algorithm \ref{alg:cppo} describes the Continual PPO algorithm.

\begin{algorithm}
  \caption{Continual PPO}
  \label{alg:cppo}
\textbf{Initialize:} All the required parameters for PPO \\
\textbf{Initialize:} All the required parameters for continual backpropagation\\
\For{iteration = 1, 2 ... }{
    \textbf{Collect data:} Run the current policy to collect a set of trajectories \\
    \For{epochs = 1, 2 ... }{
    Divide and shuffle the collected trajectories into mini-batches\\
    \For{each mini-batch}{
        Compute the objectives for policy and value networks\\
        Update the weights of both networks using Adam\\
        Update the weights of both networks using selective reinitialization\\
    }
}
}
\end{algorithm}

\begin{algorithm}
  \caption{Continual backpropagation (CBP) with Adam for a feed forward neural network with $L$ hidden layers}
  \label{alg:cbp-adam}
\textbf{Set:} step-size $\alpha$, replacement rate $\rho$, decay rate $\eta$, and maturity threshold $m$ (e.g. $10^{-4}$, $10^{-4}$, $0.99$, and $1000$) \\
\textbf{Initialize:} Set moment estimates $\beta_1$, and $\beta_2$ (e.g., $0.9$, and $0.99$) \\
\textbf{Initialize:} Randomly initialize the weights $\textbf{w}_0 , ...,\textbf{w}_L$. Let, $\textbf{w}_l$ be sampled from a distribution $d_l$\ \\
\textbf{Initialize:} Utilities $\textbf{u}_1 , ...,\textbf{u}_L$, average feature activation $\textbf{f}_1, ...,\textbf{f}_l$, and ages $\textbf{a}_1, ...,\textbf{a}_L$  to 0\\
\textbf{* Initialize:} Moment vectors $\textbf{m}_1 , ...,\textbf{m}_L$, and $\textbf{v}_1 , ...,\textbf{v}_L$ to $0$, where $\textbf{m}_l$, $\textbf{v}_l$ are the first and second moment vectors for weights in layer $l$\\
\textbf{* Initialize:} timestep vectors $\textbf{t}_1 , ...,\textbf{t}_L$ to $0$, where $\textbf{t}_l$ is the timestep vector for weights in layer $l$\\
\For{each input $x_{t}$}{
    \textbf{Forward pass:} pass input through the network, get the prediction, $\hat y_t$ \\
    \textbf{Evaluate:} Receive loss $l(x_t, \hat{y}_t)$ \\
    \textbf{Backward pass:} update the weights using stochastic gradient descent \\
    \textbf{* Update timestep:} Increase $\textbf{t}_1 , ...,\textbf{t}_L$ by $1$ \\
    \textbf{* Update moment estimates:} Update $\textbf{m}_1 , ...,\textbf{m}_L$, and $\textbf{v}_1 , ...,\textbf{v}_L$, using the newly computed gradients \\
    \For{layer $l$ in $1:L$}{
        \textbf{Update age:} $\textbf{a}_l\: +\!= 1$ \\
        \textbf{Update feature utility:} Using Equation 5\\
        \textbf{Find eligible features:} Features with age more than $m$\\
        \textbf{Features to replace:} $n_l\!*\!\rho$ of eligible features with smallest utility, let their indices be $\textbf{r}$ \\
      \textbf{Initialize input weights:} Reset input weights $\textbf{w}_{l-1}[:, \textbf{r}]$ using random samples from $d_{l}$\\
      \textbf{Initialize output weights:} Set $\textbf{w}_{l}[\textbf{r}, :]$ to zero\\
      \textbf{Initialize utility, feature activation, and age:} Set $\textbf{u}_{l, \textbf{r}, t}$, $\textbf{f}_{l, \textbf{r}, t}$, and $\textbf{a}_{l, \textbf{r}, t}$ to 0 \\

    \textbf{* Initialize moment estimates:} Set $\textbf{m}_{l-1}[:, \textbf{r}]$, $\textbf{m}_{l}[\textbf{r}, :]$, $\textbf{v}_{l-1}[:, \textbf{r}]$, and $\textbf{v}_{l}[\textbf{r}, :]$ to $0$ \\
    \textbf{* Initialize timestep:} Set $\textbf{t}_{l-1}[:, \textbf{r}]$, and $\textbf{t}_{l}[\textbf{r}, :]$ to $0$ \\
}
}
\hrulefill\\
* These are Adam specific updates in continual backpropagation \\
The inner for-loop specifies selective reinitialization based updates
\end{algorithm}

\end{document}